\RequirePackage{fix-cm}
\documentclass[smallextended,final]{svjour3}       
\smartqed  
\usepackage{times}
\usepackage{hyperref}
\usepackage{url}
\usepackage{graphicx,tabularx}
\usepackage{amsmath}
\usepackage{amsfonts}
\usepackage{graphicx}
\usepackage{cite}

\usepackage{amssymb}
\usepackage{subcaption}
\usepackage{comment}
\usepackage{algorithm}
\usepackage{algorithmic}
\usepackage{caption}
\usepackage{wrapfig}
\usepackage{subcaption}
\usepackage{booktabs,tabularx}
\usepackage{wrapfig}
\usepackage{multirow}
\def\eg{{e.g. }}

\def\ie{{i.e. }}
\newcommand{\ignore}[1]{}
\usepackage{listings}  	
\usepackage{amssymb}
\usepackage[toc,page]{appendix}

\let\emptyset\varnothing

\AtBeginDocument{%
  \setlength{\oddsidemargin}{\dimexpr(\paperwidth-\textwidth)/2-1in}%
  \setlength{\evensidemargin}{\oddsidemargin}%
  \setlength{\topmargin}{%
    \dimexpr(\paperheight-\textheight)/2-\headheight-\headsep-1in}%
}
\begin{document}

\title{Text to Multi-level MindMaps: A Novel Method for Hierarchical Visual Abstraction  of Natural Language Text}



\author{Mohamed Elhoseiny         \and
        Ahmed Elgammal 
}

\authorrunning{M Elhoseiny et al.}

\institute{Mohamed Elhoseiny, Ahmed Elgammal \at
	       Department of Computer Science, Rutgers University\\
              110 Frelinghuysen Road, Piscataway, NJ 08854-8019, USA \\
              \email{m.elhoseiny@cs.rutgers.edu, elgammal@cs.rutgers.edu}           
}

\date{Received: date / Accepted: date}

\maketitle
\begin{abstract}
MindMapping \cite{mmapbook:buzan} is a well-known technique used in note taking, which encourages learning and studying. MindMapping has been manually adopted to help present knowledge and concepts in a visual form. Unfortunately, there is no reliable automated approach to generate MindMaps from Natural Language text. This work firstly introduces MindMap Multilevel Visualization concept which is to jointly visualize and summarize textual information. The visualization is achieved pictorially across multiple levels using semantic information (i.e. ontology), while the summarization is achieved by the information in the highest levels as they represent abstract information in the text. This work also presents the first  automated approach that takes a text input and generates a MindMap visualization out of it. The approach could visualize text documents in multilevel MindMaps, in which a high-level MindMap node could be expanded into child MindMaps. \ignore{ As far as we know, this is the first work that view MindMapping as a new approach to jointly summarize and visualize textual information.} The  proposed method involves understanding of the input text and converting it into intermediate Detailed Meaning Representation (DMR). The DMR is then visualized with two modes; Single level or Multiple levels, which is convenient for larger text. The generated MindMaps from both approaches were evaluated based on Human  Subject experiments performed on Amazon Mechanical Turk with various  parameter settings.

\end{abstract}

\section{Introduction}
\label{sec:1}
\begin{figure}[ht!] 
 \vspace{-5mm}
 \centering
    \includegraphics[width=1.0\textwidth]{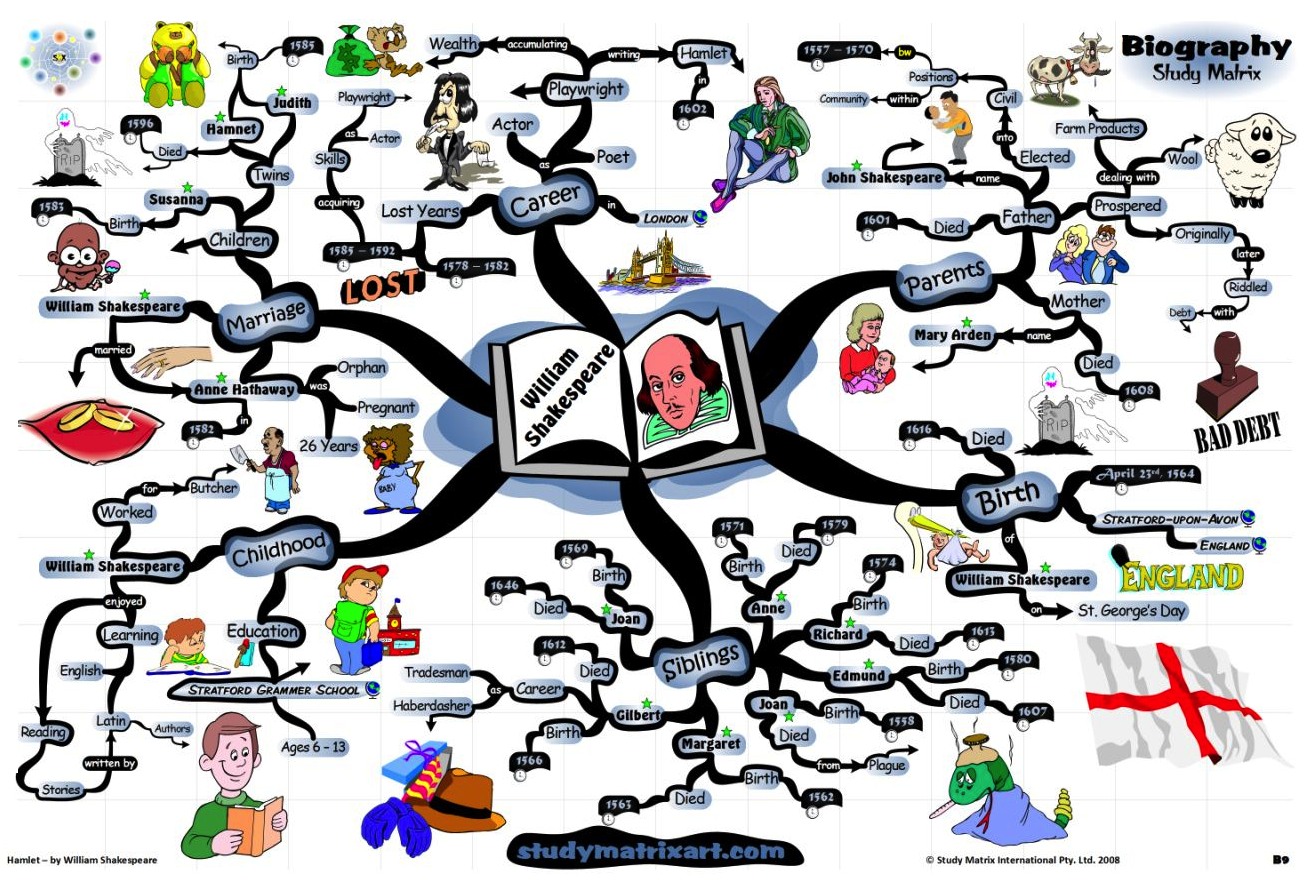}
  \vspace{-5mm}
     \caption{Manually Created Shakespeare's life MindMap~\cite{ShMMURL:2014} (best seen electronically)}
 \label{fig:MMMMapEx}
\end{figure}

{M}{indMapping} was introduced as a visual note taking technique, developed by Tony Buzan in the 1960s \cite{mmapbook:buzan}. It is a powerful pictorial method for representing knowledge, concepts and ideas. \ignore{Figure \ref{fig:MMapEx} represents linear text of Shakespeare's life, and its corresponding manually drawn MindMap. }Figure~\ref{fig:MMMMapEx} shows a MindMap example that illustrates William Shakespeare's biography. It is not hard to see that by just looking at this picture, it is easy to recall many aspects of Shakespeare's life, which is one of the MindMap benefits (\ie Review, Revision).  Creating MindMaps itself is also interesting compared with linear text. It also improves imagination and meanwhile it is fun. 

``Linear note taking has served as one of the greatest impediments to learning'', says Tony Buzan. However, he claimed that linear note taking constraints the brain's potentials by forcing it to use limited modes of expression \cite{mmapbook:buzan} and hence it fails to stimulate creativity. \ignore{While MindMaps were designed to simulate both of the brain's cerebral hemispheres, which encourages memory.} According to Buzan~\cite{mmapbook:buzan}, the main motivation behind MindMapping technique is that, unlike linear text, using pictorial representation of ideas stimulates both sides of the brain (left and right). However, there are scientific doubts about Buzan's argument on the brain hemispheres. Recently, neurologists at Utah university performed a two-year study on 1000 people \cite{leftrightEval13}, which  shows that the brain is more complex than this claim of left and right-sided brain. They divided the brain into 7000 regions and they found no evidence that the human subjects had a stronger right or left-sided brain network \footnote{More discussion on this study is covered by the guardian \text{http://www.theguardian.com/commentisfree/2013/nov/16/left-right-brain} \text{-distinction-myth}}.  Despite that Buzan's justification of MindMaps by brain hemispheres is mere popular psychology and not scientifically justified, there are  a lot of studies that shows that the MindMapping is a powerful and effective technique to access   information and to  encourage learning, studying and reading (\eg~\cite{mmapstudy1,mmapstudy2}).

With today's vast and rapid increase in the use of electronic document readers, smart phone, and tablets,  there is an eminent need to develop methods  for visualizing and summarizing textual contents. We argue that MindMapping is not only a note-taking tool, but also can be developed to serve as a  powerful tool for joint text summarization and visualization. Converting a text paragraph to a MindMap would provide an easier way to visually access the knowledge and ideas in the text. 

\begin{figure}[hb!]
 \vspace{-7mm}
   \centering
   \includegraphics[width=0.8\textwidth]{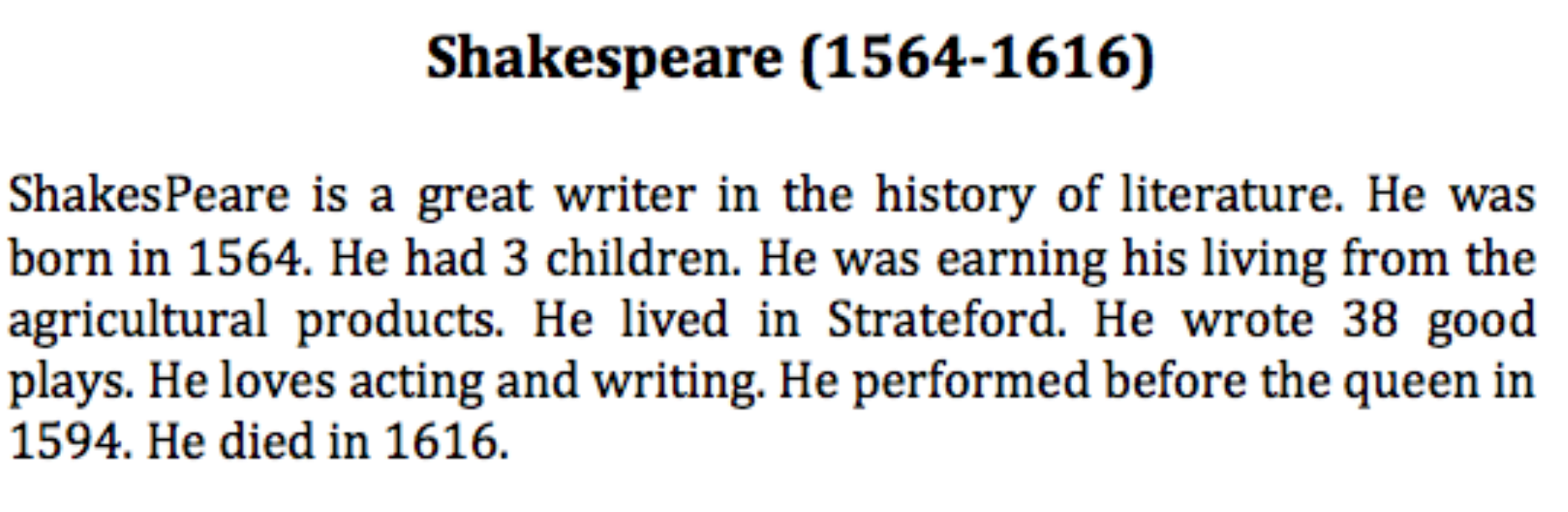}
    \includegraphics[width=0.7\textwidth,totalheight = 0.21\textwidth]{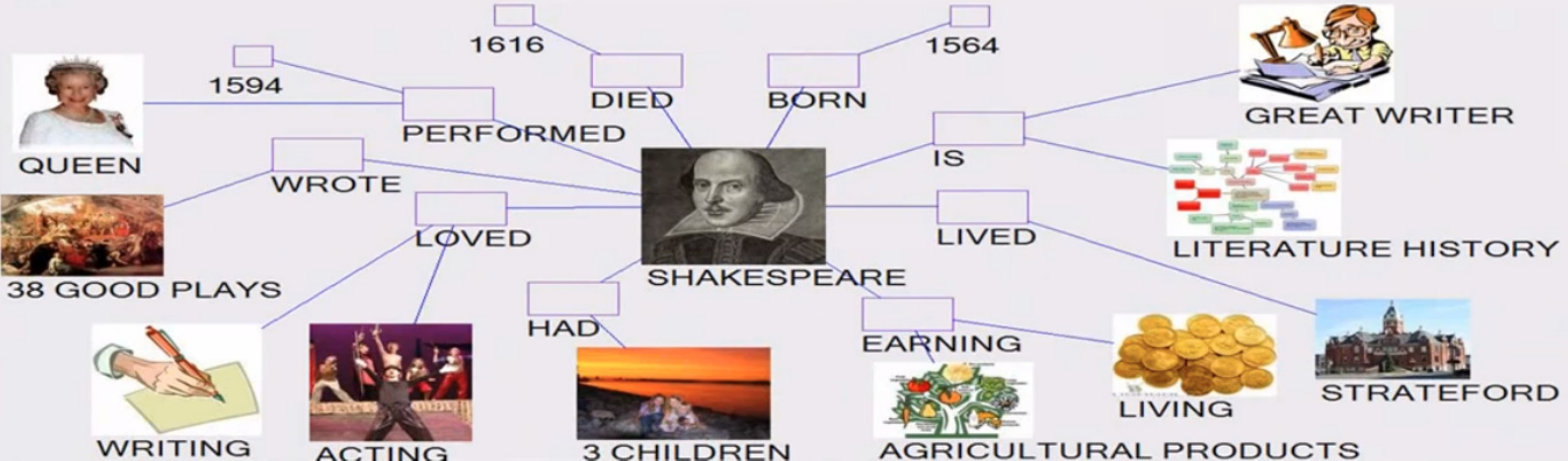}
    \includegraphics[width=0.95\textwidth,totalheight = 0.44\textwidth]{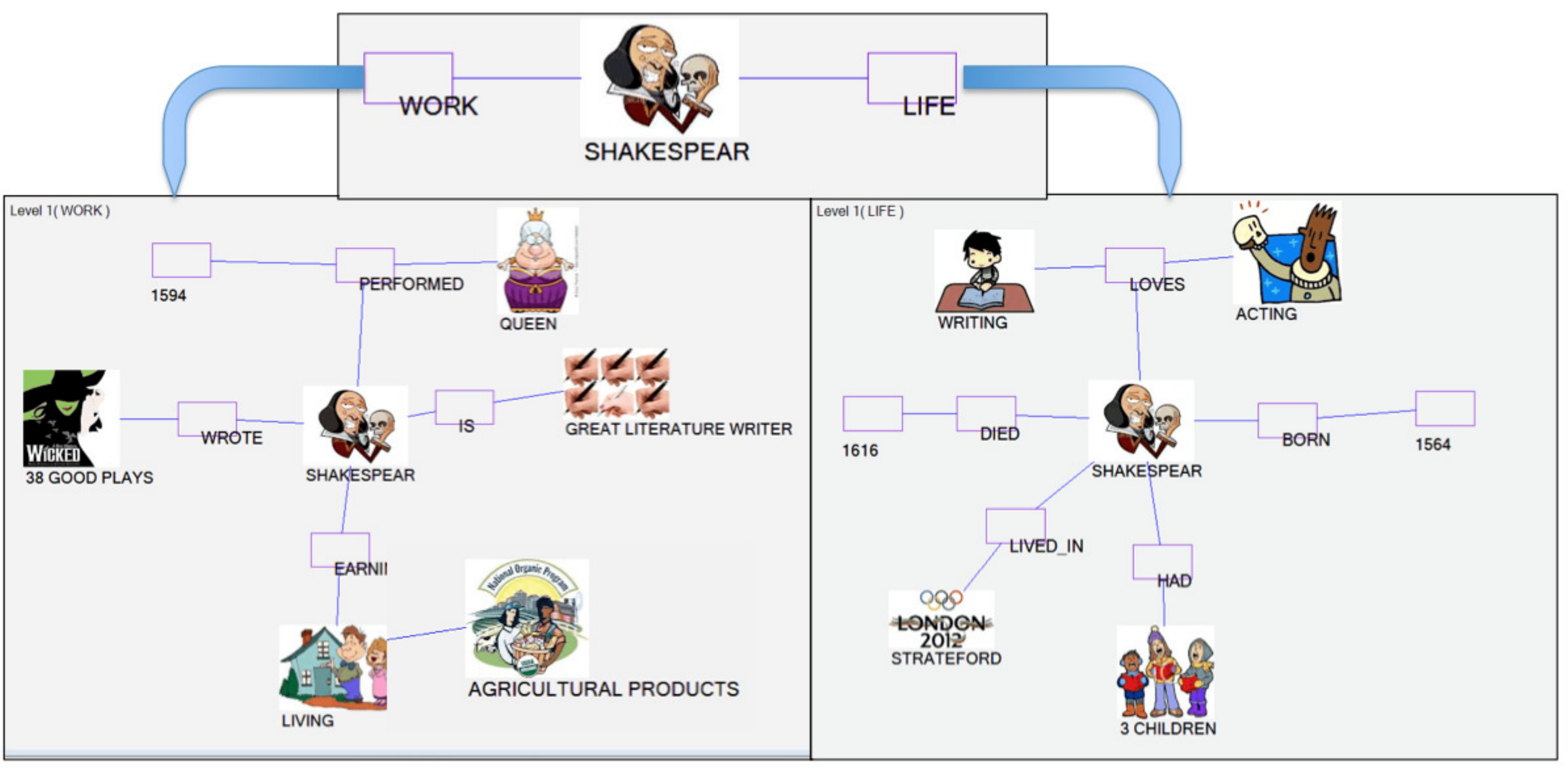}
     \vspace{-3mm}
 \caption{Top Left: Input text, Top Right: Shakespeare Single-Level (With web image queries has no restriction on images),  and Bottom: Multi-Level MindMaps (with web image queries restricted to the set of Clip-Art Images), generated by our Framework and their corresponding input textual description}
 \label{fig:shakessystem}
\end{figure}

One of the reasons why many people may not create  MindMaps is that it needs a huge mental effort and  concentration, specially for MindMapping a big text. Besides, only few people are creative enough to draw good MindMaps. \ignore{To the best of our knowledge, there is no practical approach available to automatically convert text documents into a MindMap representation. Therefore, there is no industrial tools released to do this task. One possible justification is  inapplicability of the current methods to visualize large text. } To the best of our knowledge, there is no existing method to build MindMaps from raw text according to the joint text summarization and visualization concept that we develop in this paper. The goal of this work is to build a step in that direction by developing the first framework that takes a text input and generates a MindMap visualization out of it;  a former version was presented in \cite{mmism12}. Our framework is constructed to visualize large text documents in  multilevel MindMap, in which a high-level MindMap node could be expanded into a child MindMap. Figure~\ref{fig:shakessystem} shows the MindMap visualization of the Shakespeare example, generated by our method, in a single level and multiple levels ( his life and work information are expanded as two child MindMaps).

There are various potential applications for these  versions of automatically created MindMaps that we are developing. For example, audio recording of notes can be converted into text that is passed into our framework to generate a MindMap notes. A lecturer can utilize the framework to prepare presentations by converting text to MindMaps. Furthermore, the generated hierarchical MindMaps would help make MindMap presentations more common due to minor effort and time needed to create them. Finally, MindMaps visualization could help users of electronic book readers to better access information through them\ignore{, which can be very beneficial}.

\noindent {\em The contributions of this paper are:}

\begin{enumerate}
	\item{We introduce MindMapping as an approach to jointly visualize and summarize textual information. Visualization is achieved pictorially across multiple levels using semantic information (i.e. ontology), while summarization is achieved by the information in the highest levels. Our work could also be viewed as a  method to visually abstract textual information in multiple levels.}
	\ignore{
      \item{Novel Architecture to generate MindMaps from pure text.}}
       \item{ Novel approach to generate Single and Multilevel MindMaps from pure text and an input Ontology.}
      \item{ Two approaches were proposed for retrieving relevant pictures from the web.}
      \item{ First comprehensive evaluation of  an automated MindMap framework by human subjects. }
\end{enumerate}

 The rest of this paper is organized as follows.  Section~\ref{sec:2} presents the related literature. Section~\ref{sec:3} to section~\ref{sec:6} describe the proposed framework and techniques. Section~\ref{sec:7}  describes the evaluation procedures and shows the  results using different system parameters. Section~\ref{sec:8}  presents our conclusion and the future work.

\section{Related Work}
\label{sec:2}
Our formulation addresses the problem as a hybrid between text summarization and visualization. In the context of text summarization, the top level MindMap serves as a summarized version/ abstraction of the document. Meanwhile, the detailed information of textual content could be visually accessed through a hierarchy of MindMaps. This section presents related work in each of  text summarization, visualization, a connection to Concept/Topic Maps, and MindMapping  existing systems.

\subsection{Text Summarization}

\ignore{Several approaches have been investigated for text summarization over the past five decades \cite{Das_Martins_2007}. By definition, text summarization takes input from a single document or multiple documents and generates a few set of statements that highlights the important information in the input document(s).}

Several approaches have been proposed for text summarization\cite{Das_Martins_2007}; key approaches follow. Conroy and O'leary \cite{Conroy:2001} used Hidden Markov Models (HMMs) to model the problem by sentence extraction/selection from a document considering local dependencies between sentences. Later, Osborne \cite{Osborne:2002} used log-linear models to obviate the feature independence assumption in previous models as he claimed and showed empirically that the system produced better extracts than a naive-Bayes model, with a prior appended to both models. Then, Kaikhah et al \cite{txtsummNN04} trained a Neural Network model from the labels and the features for each sentence of an article, that could infer the proper ranking of sentences in a test document. Other approaches use deep Natural Language Processing (\eg  \cite{Marcu:1998}, where  Marcu et al merged the discourse based heuristics with traditional heuristics).

While, text summarization provides a reasonable resolution to the growing information overload, it does not provide an organized access to the document information. This limits its capability to structurally localize the information of user's interest in the given document, which introduces a key motivation for our multilevel MindMap Visualization concept . 

\subsection{Text Visualization}

\ignore{There are few relevant approaches for text visualization.} Some of the state-of-the-art text visualization techniques build on tag clouds. Examples include work conducted by IBM in Wordle, which is a web-based text visualization method that creates a tag-cloud-like displays with careful attention to typography, color and decomposition~\cite{Viegas09}. The generated clouds give greater prominence to \ignore{\begin{wrapfigure}{r}{0.39\textwidth}   
  \vspace{-3mm}
    \includegraphics[width=0.39\textwidth]{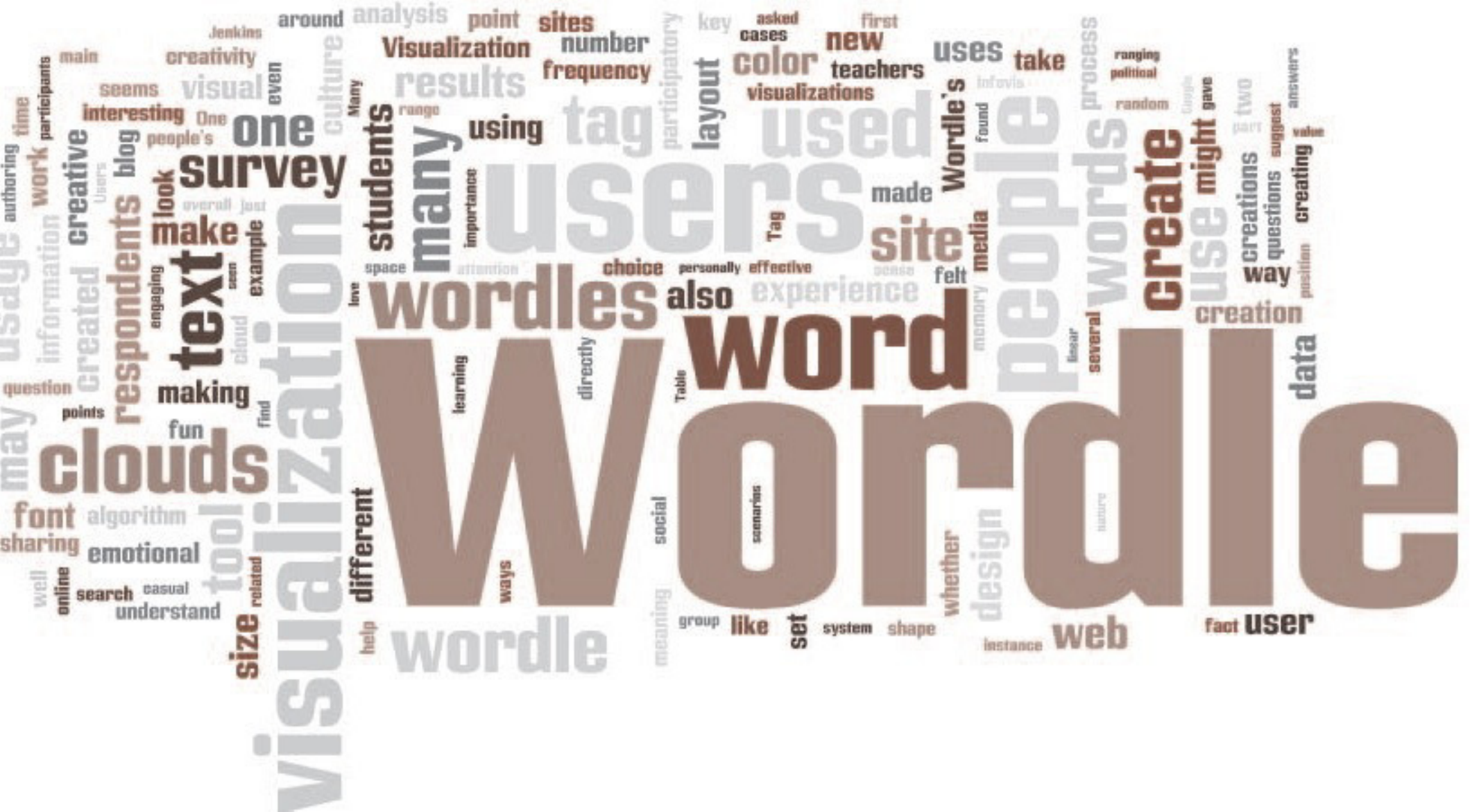} \includegraphics[width=0.39\textwidth]{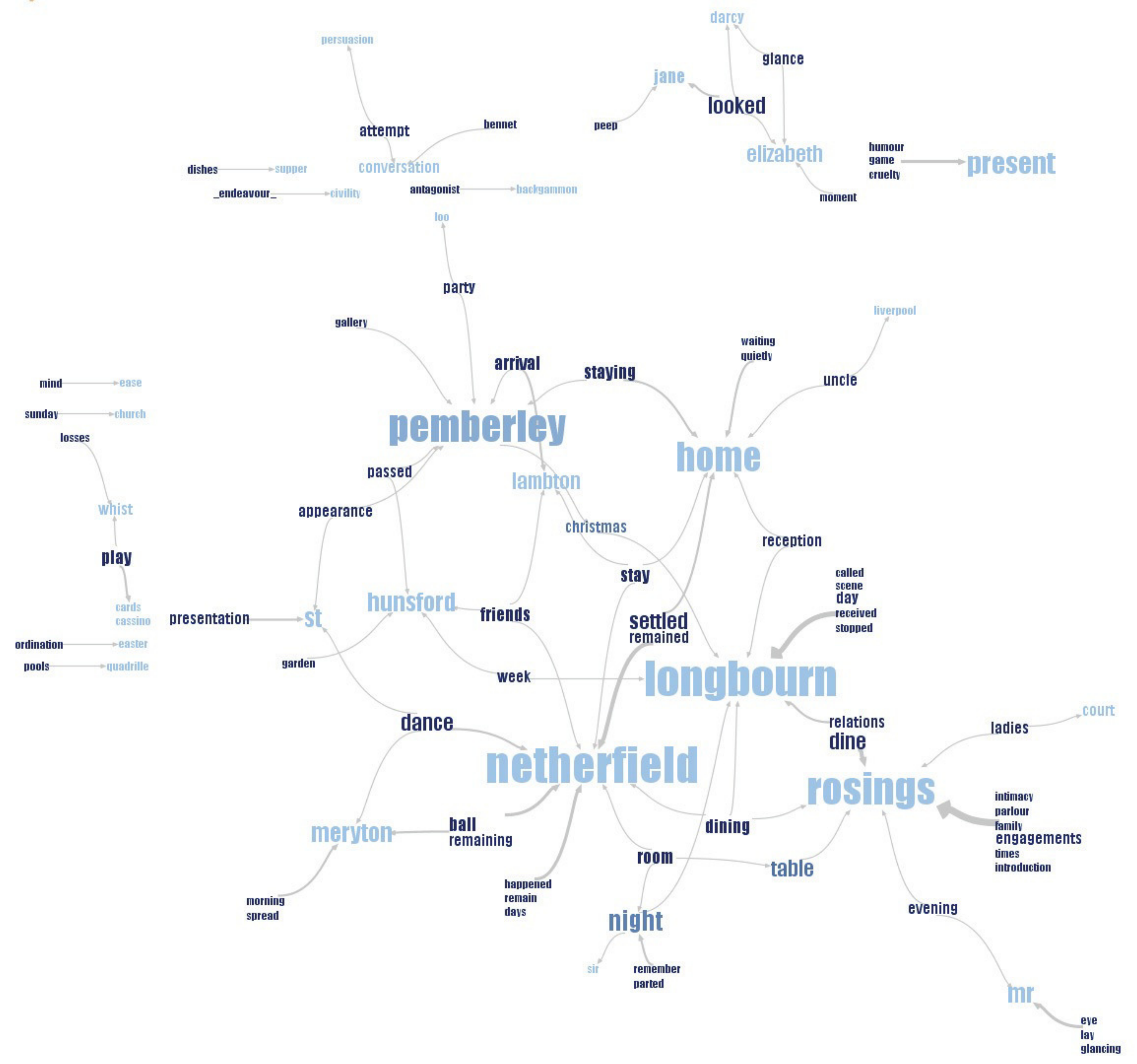}
  \caption{Wordle (top) \cite{Viegas09} and PhraseNet (bottom) \cite{vanHam:2009}}
  \label{fig:WordlePhNet}
    \vspace{-7mm}
\end{wrapfigure}} words that appear more frequently in the source text.  Van Ham et al.~\cite{vanHam:2009} presented  PhraseNet technique,  which diagrams the relationships between different words in the text. It uses a simple form of pattern matching to provide multiple views of the concepts contained in a book, speech, or poem. Recently, Afzal et al~\cite{Afzal_TVCG12} integrated spatial constraints to  automatically visualize  textual information on  typograhical maps. \ignore{Wordle PharseNet The aforementioned approaches  might look similar to our problem (see Figure~\ref{fig:WordlePhNet}. However,} There are  three significant differences differentiate our work from the aforementioned works. First, their algorithms do not  involve deep understanding of the text\ignore{, so the building block in the tag-cloud methods is a single word}. Second, each node is visualized with the word itself. However in MindMaps, visual nodes could be mapped into pictures of a concept and its properties (\eg car with red color attribute). Third, our approach can generate multilevel representation, which is not supported by  Wordle,  PhraseNet or Afzal's method. 

\begin{figure}[h!]
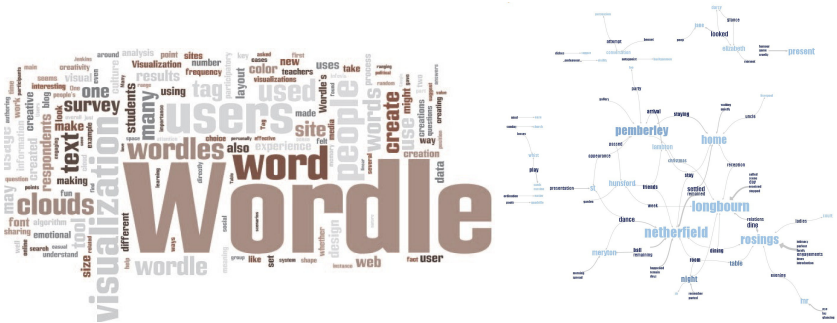

\vspace{-5mm}
  \includegraphics[width=0.55\textwidth]{wordle.eps}
    \includegraphics[width=0.38\textwidth]{phnet.eps}
    \vspace{-3mm}
  \caption{Wordle (left) \cite{Viegas09} and PhraseNet (right) \cite{vanHam:2009}.}
 \label{fig:WordlePhNet}
     \vspace{-3mm}
\end{figure}

Topic models and semantic representations have been adopted to help organize and visualize information.  In this direction, Feng et al.~\cite{Feng_Lapata_2010} developed a model for  multimodal meaning representation, based on linguistic and visual context. The approach exploits  enormous resource of documents and associated images available on the web. A  relevant application is the automatic conversion of text to 3D animation. Ma~\cite{Ma06} proposed Lexical Visual Semantic Representation (LVSR) which connects linguistic semantics to the visual semantics and is suitable for action execution. This application involves deep analysis of the text converting it into user defined concept by which it converts the text into actions to be performed on 3D models. In addition, Coyne et al.~\cite{Coyne:2010} presented a method that incorporates lexical and real-world knowledge to depict scenes from language. However, these approaches  are constrained to limited number of predesigned 3D-models and relations and are not hierarchical. Recently, Dou et al~\cite{Wenwen_TVCG113} presented an approach to visualize text collections using topic hierarchies. While,  their approach provides a reasonable access for visually exporing text corpus,  our approach addresses pictorial visualization and abstraction of a chosen textual document by the MindMap Visualization concept.  In addition, the explored hierarchy in~\cite{Wenwen_TVCG113} are topic based, while the hierarchy provided by our visualization is based on semantic abstraction of the textual input content.  As an illustration, our approach could be integrated with the  approach~\cite{Wenwen_TVCG113} to visualize a document selected by exploring the topic hierarchies,  organized by~\cite{Wenwen_TVCG113}.

\subsection{Topic Maps and Concept Maps}

\ignore{We begin our discussion about this connection by defining Concept Maps and Topic maps.} \textit{Concept maps}~\cite{Novak06thetheory} have been used to express people's understanding about a specific topic. For few decades, this tool has attracted people of all ages and different knowledge domains. They include concepts usually enclosed as boxes or circles and relationships between these concepts indicated by a connecting line. \textit{Topic maps}~\cite{TMap2003,TMap2006} are very similar to Concept Maps, while Topic Maps are standardized. It has been defined as a form of semantic web technology and a lot of work has been done to address interoperability between the W3C's RDF/OWL/SPARQL. The current standard is taking place  under the umbrella of the W3C committee. \ignore{The current standard is taking place  under the umbrella of the ISO/IEC JTC1/SC34/WG3 committee \cite{ISO1,ISO2}.}

In contrast to Concept Maps and Topic Maps, MindMapping \ignore{of an idea }often includes but is not limited to radial hierarchies (where central ideas is in the center, while connected to related entities around it) and tree structures. Another difference is  that a MindMap reflects people's thoughts about a single topic, while Concept Maps/ Topic Maps can present a real or abstract system  or a set of concepts, which is not restricted to specific topic. 

The majority of the literature has focused on the representation of topic maps and concept maps to enhance some NLP tasks (e.g.  parsing, Word Sense Disambiguation \cite{WSDSurvey09}, etc). Rather, we would like to view these maps as a powerful tool that could be used to jointly visualize and summarize information. In particular, we present the concept of multilevel/hierarchical MindMaps as a notion of visualizing textual information  in multiple levels. Higher levels reflects coarse information in the text document, while lower levels present the fine details. We chose to name it multilevel/hierarchical MindMaps because MindMaps are (1) less restrictive on the representation than Concept and Topic Maps, (2) topic-oriented which is relevant in the setting we utilize (i.e. representing information from textual description), and (3) In our experiments, we focus on topics that could be presented as a single central idea, which is consistent with the MindMap notion. However, we argue that most of the topics could be presented a set of central ideas that could be presented by MindMaps. Hence, we don't restrict our definition of Hierarchical MindMaps to topics of a single central idea. We focus in this paper in the steps, we designed to generate the multilevel MindMap from a textual description.

\subsection{Existing MindMapping Systems}

There are two ways for  creating a MindMap. One way is  manually by using a paper, pens and colors, but this requires, beforehand, reading and understanding the \ignore{
\begin{wrapfigure}{r}{0.39\textwidth}   
  \vspace{-8mm}
        \includegraphics[width=0.38\textwidth]{mmmanual.eps}
 \caption{The process of Manual MindMap Generation}
 \label{fig:MMapManual}
    \vspace{-7mm}
\end{wrapfigure} }text well to design a good MindMap. Another way is to use  existing software that just serves as an editing canvas, where user can insert pictures and create relations between them (i.e., manually). Examples include ``I Mind''~\cite{IMindTool:2012}, ``Nova Mind''~\cite{NovaMindTool:2012}\ignore{; Figure~\ref{fig:MMapManual} illustrates the process of MindMap manual generation}.

Finally, the most relevant work was conducted by Hamdy et al. \cite{MM09}. They presented a prototype for the MindMap Automation, however, it was just a demonstration on few examples of few sentences and then it was applied in a mobile application in \cite{MMMobile11}. There are  two critical drawbacks in this approach. (1) It is almost an unevaluated demo. (2) It supports only single-level MindMaps. This limitation make the approach incapable of representing information of a larger text. Hence, it can not support our concept of simultaneously visualizing and summarizing textual information.

\section{Proposed Framework}
\label{sec:3}

\begin{figure}[ht!]
\vspace{-5mm}
   \centering
    \includegraphics[width=1.0\textwidth]{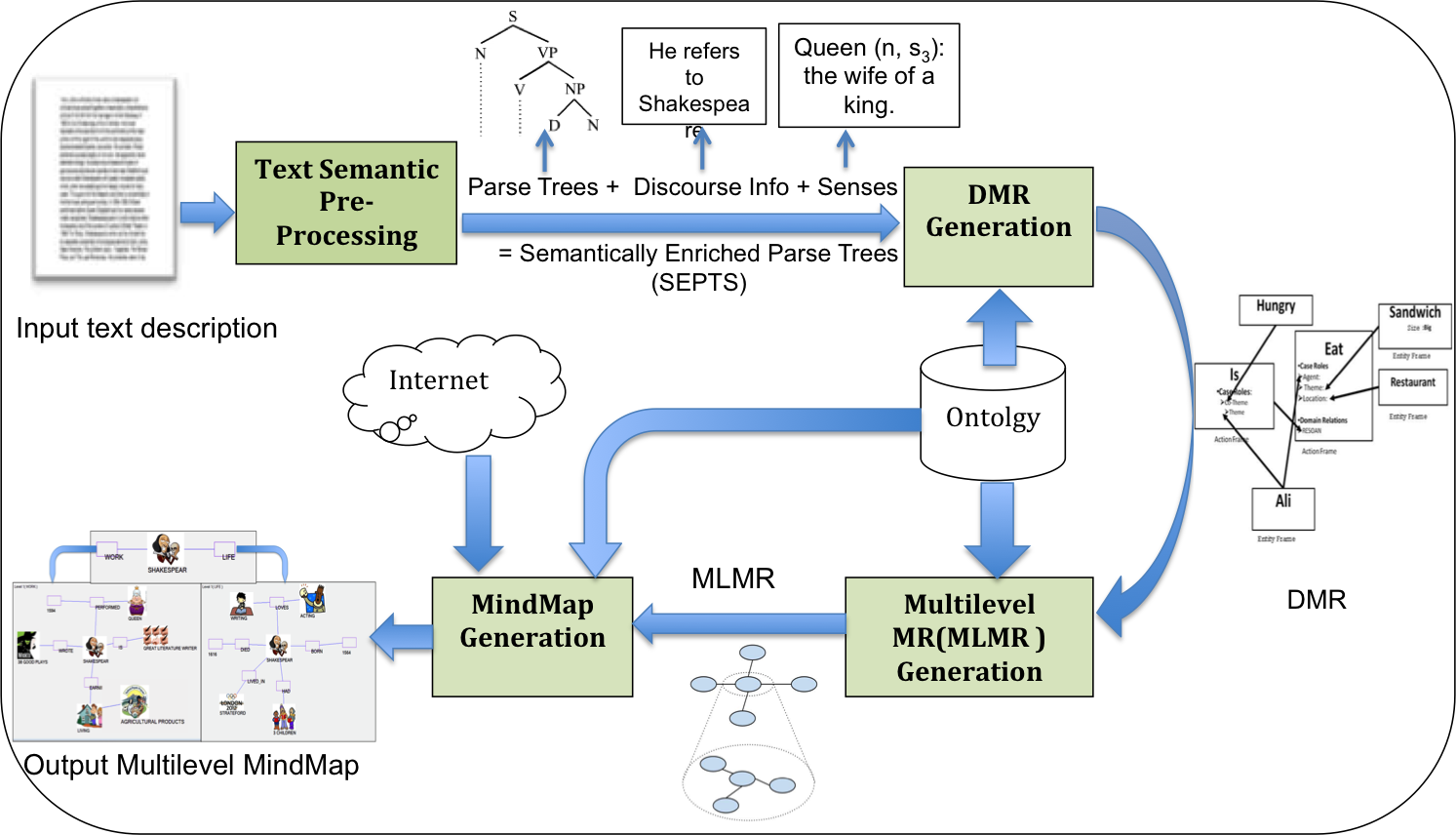}
  \caption{Data Flow  of the proposed Multilevel MindMap Generation framework}
   \label{fig:MMSysArch}
   \vspace{-5mm}
\end{figure}

Our objective is to propose a method to jointly visualize and summarize visual information (we call multilevel MindMap visualization). We begin by presenting the outline of the approach through the data flow diagram in figure~\ref{fig:MMSysArch}; we will present the details in later sections.

\begin{figure*}[ht!]
  \vspace{-2mm}
   \centering
    \includegraphics[width=0.7\textwidth]{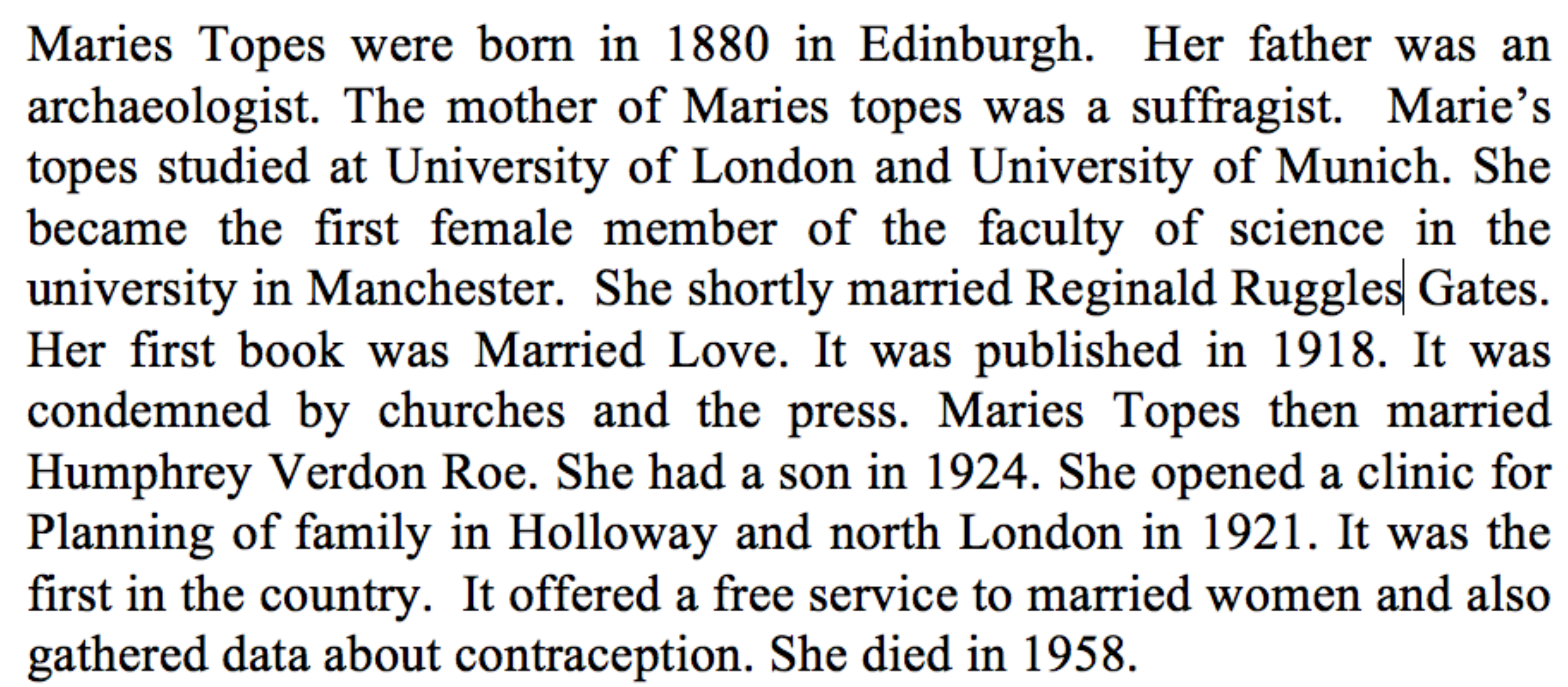}
    \includegraphics[width=0.8\textwidth]{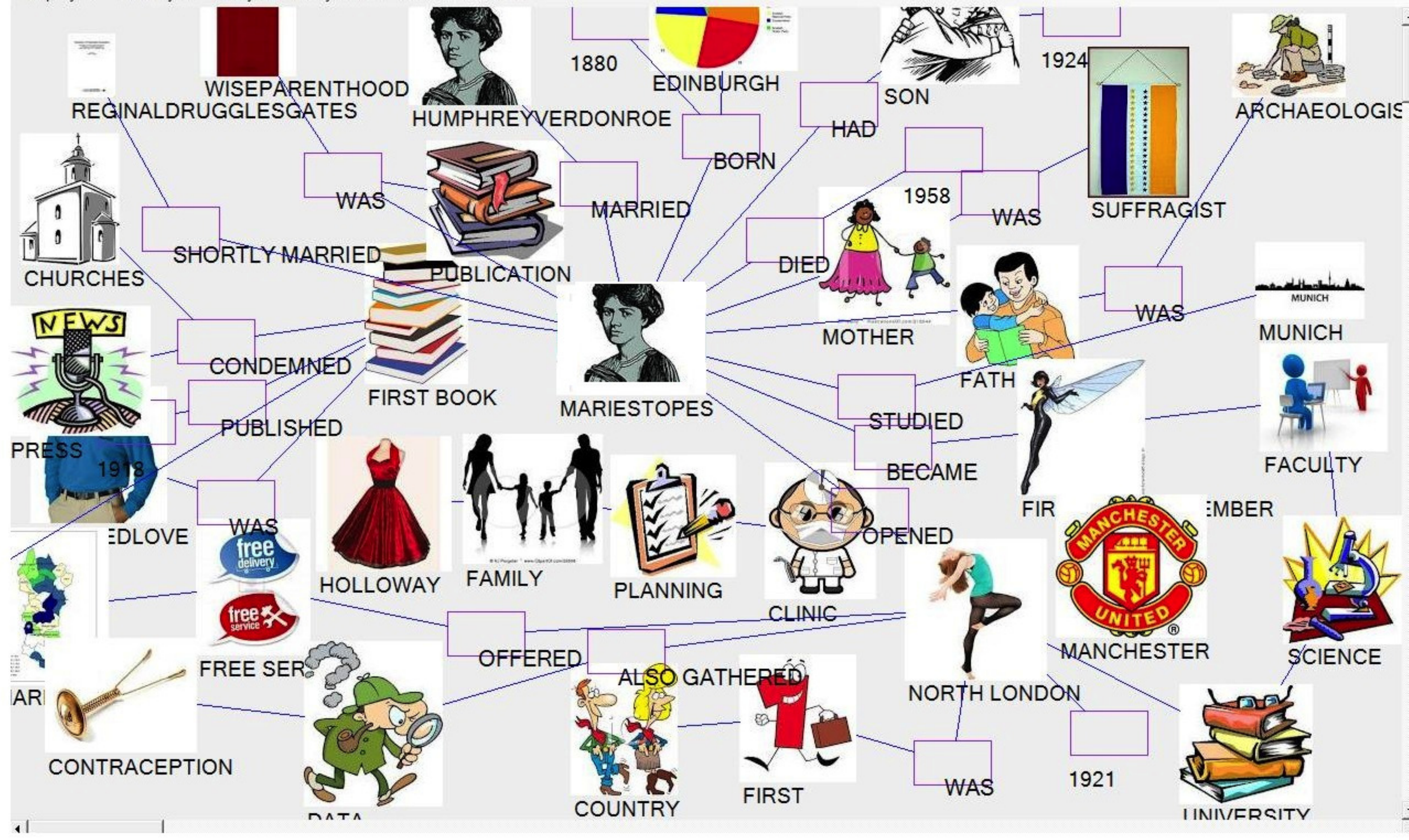}
    \includegraphics[width=0.95\textwidth]{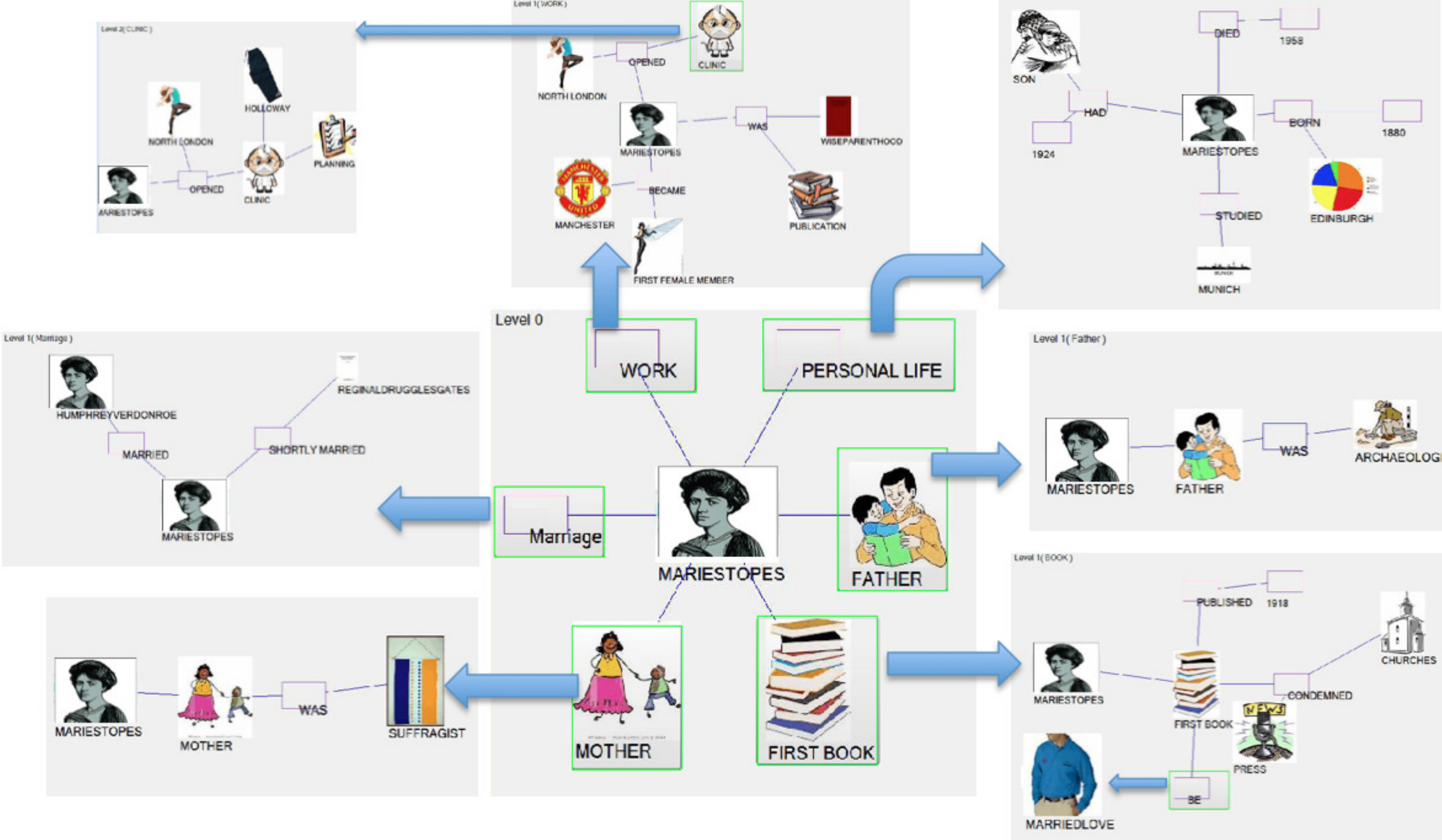}
      \vspace{-2mm}
  \caption{\textit{Top:} Textual description of "Marie Stopes" historical figure, \textit{Middle:} its corresponding single Level MindMap (Difficult to read),  \textit{Bottom:} Structured Visualization using  Multilevel MindMap Visualization of of the same text (Easier to read)}
  \label{fig:SLMMP2}
  \vspace{-5mm}
\end{figure*}

 We first  extract, what we call Semantically Enriched Parse Trees (SEPTs), from the input text through a Text Semantic Preprocessing step that we implemented\ignore{ \footnote{The text preprocessing module is gray color in figure{fig:MMSysArch} since it extracts the information from text using existing methods}}. The SEPTs consist of the parse trees for each sentence augmented with disambiguated senses at each terminal node in the parse tree. Furthermore, for each anaphor node (\eg pronouns), a pointer, to what it refers to, is stored.  These SEPTs are then converted into a Detailed Meaning Representation (DMR) that represent all the information in the text as a graph. Then, the DMR is converted to a Multi-Level Meaning Representation (MLMR), which is a tree of meaning representations. Each node in the MLMR tree stores a Meaning Representation (MR) starting from the root node where its MR holds the most abstract information in the text to the leaves where details are presented. The final step in our framework is the MindMap Generation, which converts an input MR into its pictorial representation \ignore{(\ie               see figure~\ref{fig:MMapEx}) }and dynamically allocate it on the screen. Hence, the MLMR could be visualized as a Multilevel MindMap by traversing the MLMR tree and visualizing the MR of each node. It is worth mentioning that a Single level MindMap could be generated by running the MindMap Generation step directly on the DMR instead of Generating a MindMap for each node in the MLMR. However, If the input text is of a big size, the Single Level MindMap  would be quite unclear and unorganized. Figure~\ref{fig:SLMMP2} (middle part) shows an example, where single level Mindmap visualization hinders comprehension speed, simplicity and clarity. However, Multilevel MindMap visualization is much clearer (see figure~\ref{fig:SLMMP2} bottom part). 

An ontology is needed for both DMR Generation to understand attributes of objects/subjects  while creating the DMR, and for MLMR generation  to  semantically group related entities in MLMR. Ontology is also used in the MindMap Generation step to determine whether a concepts needs to be pictorially represented; see figure~\ref{fig:MMSysArch}.   The details of the outlined approach is presented as follows. Section~\ref{s31} details the Text Semantic Pre-processing (TSP) step. Section~\ref{sec:4} presents  the DMR generation approach. Section~\ref{sec:5} presents the MLMR Generation. Finally, section~\ref{sec:6} presents the MindMap Generation step.

\section{Text Semantic PreProcessing (TSP)}
\label{s31}

In this step, we extract sentence-wise information from plain text, which we call Semantically Enriched Parse Tree (SEPT). As aforementioned, each SEPT consists of a parse tree, anaphora resolution (discourse analysis) results, and the intended sense for each word in the sentence. In order to compute the SEPTs, four main tasks are performed: (1) morphological analysis \cite{Zhang2009}, (2) parsing \cite{stanford_dependencies,Collins:2003,Bataineh2009} and syntactic structural analysis \cite{Litkowski_2001,Lappin:1990}, which produces a single parse tree that respects the structure of given sentence. (3) discourse analysis  \cite{Qiu04apublic,Leass94analgorithm} outputs   anaphora resolution results (4) Finally, Word Sense Disambiguation (WSD) \cite{Banerjee02anadapted,citeulike:1104130,WSDSurvey09} determines the intended sense for each word in the sentence based on WordNet \cite{WordNet95}. Having augmented the parse trees with the senses produced by WSD and the anaphora resolution results, the generation of a SEPT for each sentence in the text is concluded. In this work, we utilized recent approaches for such a well-studied phase in Natural Language Processing, as detailed in \cite{MM09}. Other methods could be applied as long as they produce the SEPTs. 

Figure~\ref{fig:MMTPM} shows TSP output of Shakespeare's example. The \ignore{\begin{wrapfigure}{r}{0.60\textwidth}   
       \includegraphics[width=0.60\textwidth]{tpm.eps}
 \caption{Text Semantic Preprocessing (Syntactic Analysis, Anaphora Resolution, Word Sense Disambiguation) for Shakespeare Example}
   \label{fig:MMTPM}
    \vspace{-5mm}
\end{wrapfigure} }bottom left part shows the generated parse trees as the output of the parsing and syntactic structural analysis. The bottom left part shows the discourse analysis output which is anephora resolution. ``HE(2,1)'' means the ``HE'' word that occurs as the first word of the second statement. The listing in the figure indicates that ``Shakespeare (1,1)'', that occurred in the first word of the first statement, is referenced in the succeeding pronouns (i.e. ``He(2,1'', ``His(4,4)'', etc).  The top right part of the figure shows the meaning/sense of each word that occurs in the sentences, while grouped in the data structure for each statement. Augmenting the parse trees with the the discourse information and senses directly produces the SEPTs. We denote the set of SEPTs extracted from the text by $\mathcal{P} = \{ p_i, i=1 \to N\}$. We define $n$ be a node in any of the SEPTs, we define $n.r$  as the node that $n$ points to in case it points to a previously declared noun/noun phrase,  $n.r = NULL$ otherwise. $n.r$ could point to a node in a possibility different SEPT. We also define $n.s$ as the sense that $n$ means in case it is a terminal node,   $n.s = NULL$ otherwise. 

\begin{figure}
\centering
 \includegraphics[width=1.0\textwidth]{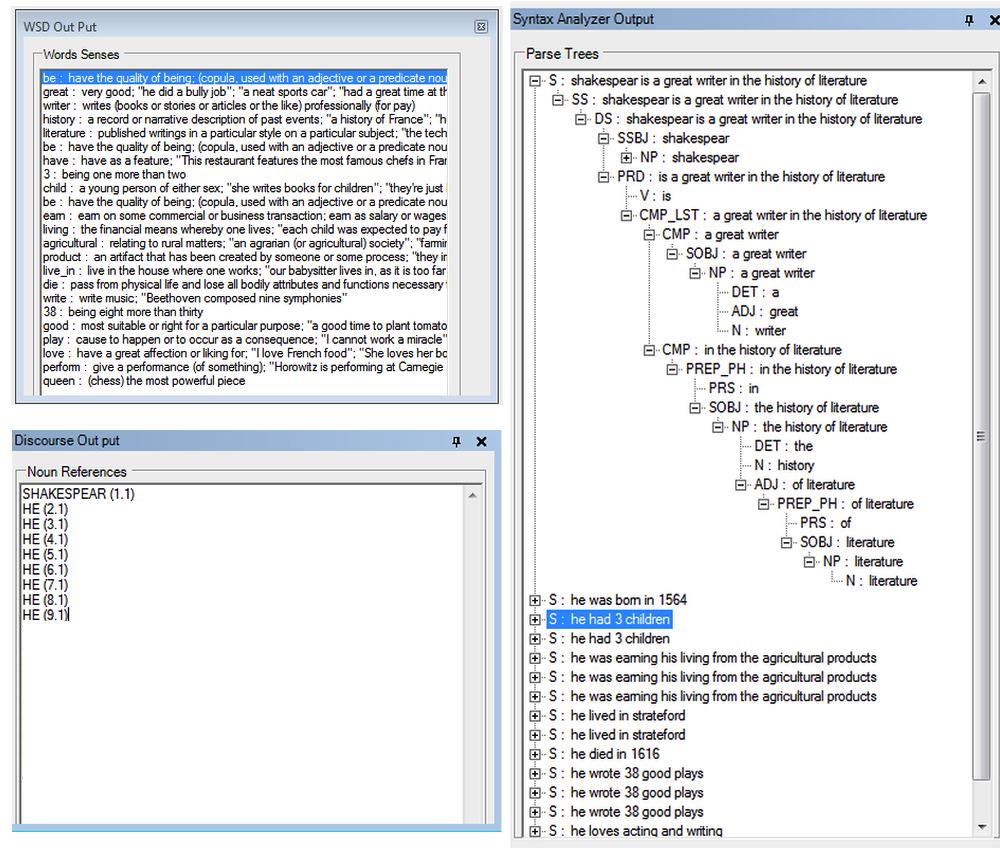}
  \caption{Text Semantic Preprocessing (Syntactic Analysis, Anaphora Resolution, Word Sense Disambiguation) for Shakespeare Example. The Syntactic Parse trees are augmented with the senses and the discourse analysis results to form SEPTs}
   \label{fig:MMTPM}
\end{figure}

\section{DMR Generation}
\label{sec:4}

\begin{figure}[h!]
\vspace{-2mm}
\centering
\begin{minipage}[c]{0.40\textwidth}
    \includegraphics[width=0.98\textwidth]{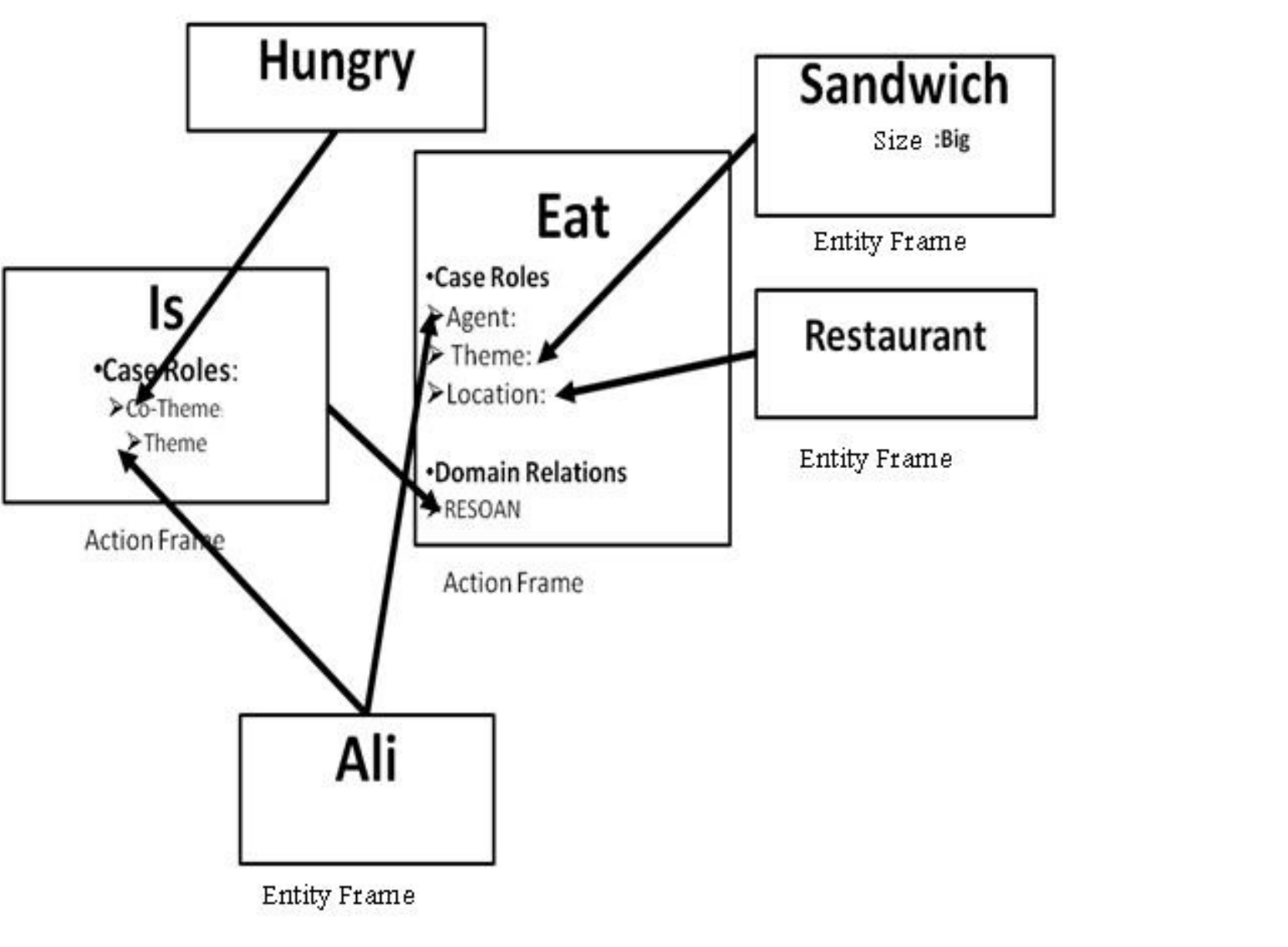}
 \caption{Example Meaning Representation for ``Ali ate the Sandwich in The restaurant because he was hungry''}
 \label{fig:MRExample}
\end{minipage}
\begin{minipage}[c]{0.04\textwidth}
\hfill
\end{minipage}
\begin{minipage}[c]{0.53\textwidth}
 \includegraphics[width=1.0\textwidth]{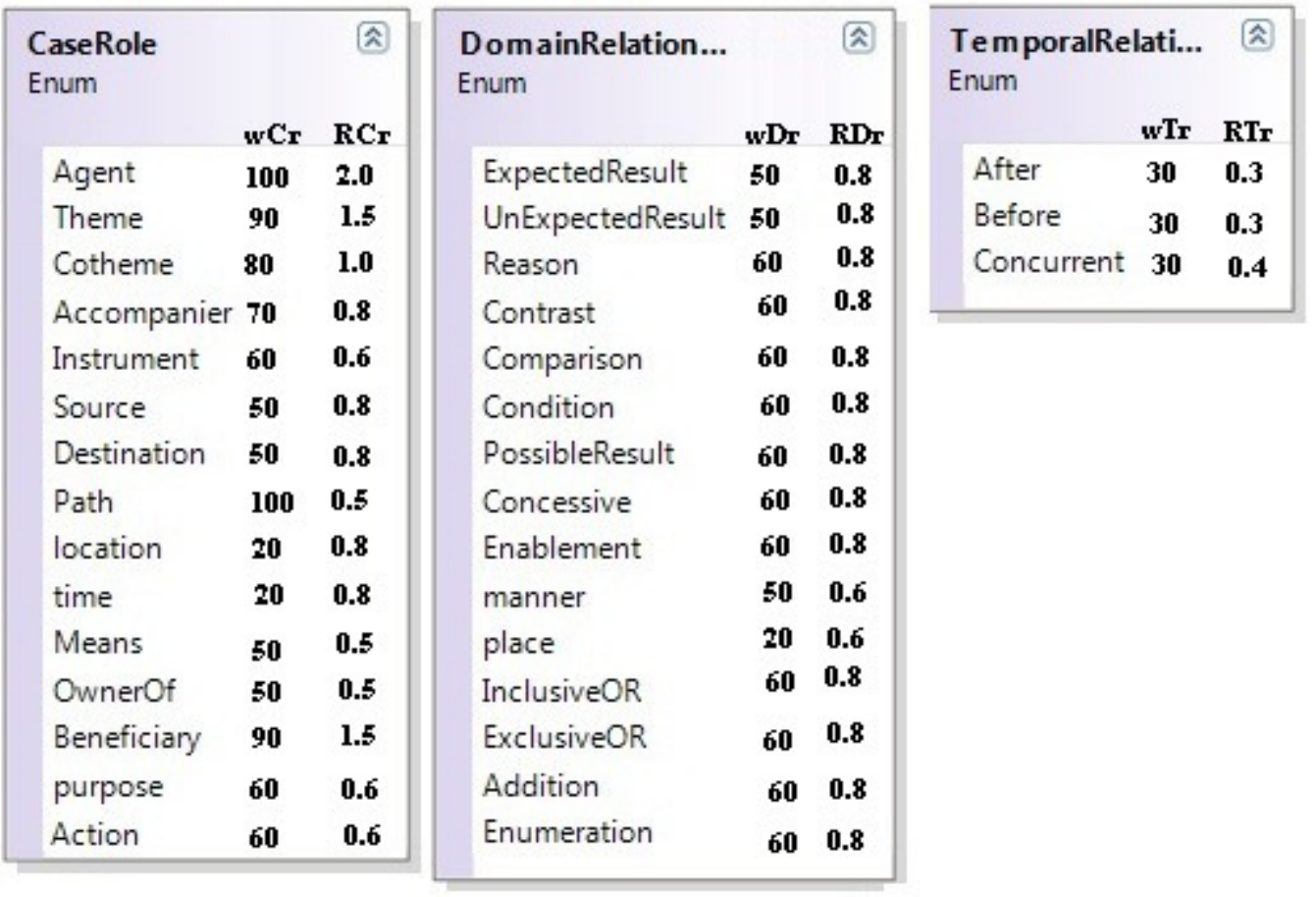}
  \caption{Relations in Meaning Representation and their weights}
  \label{fig:MRRelations}
\end{minipage}
\vspace{-5mm}
\end{figure}

Designing a meaning representation for Natural Language Text implies understanding its content and  representing it in a semantically accessible way. During DMR Generation,  $\mathcal{P}$  is  converted to a further meaningful representation, which consists of conceptual frames and relations between these them. There are  two types of frames in our framework (Entity frames and Action Frames). Entity frames represent entities mentioned in the text (persons, objects etc), while Action Frames represent activities in the text (i.e. verbs). Possible relations between the frames are Case-Role, Domain and Temporal relations.\ignore{A case-role is any of the spatially distinguished parts of an action.  } A \textit{Case-Role relation} represents semantic role of an entity frame in a particular action. For instance, in ``john plays football'' sentence, the subject ``john'' has  an ``agent''  case-role in the ``play'' action, and the object ``football'' frame has a ``theme'' case role in the same action\ignore{, where ``agent''  and ``theme''  are case-role relations}. \textit{Temporal relations} are inter-propositional relations that communicates the simultaneity or ordering in time of events or states. \textit{Domain relations} encode logical connection between action frames (i.e. verbs). Figure~\ref{fig:MRRelations} illustrates different types of case-roles, domain relations and temporal relations that are designed in our framework. Having detected the frames and the relations, the final output of the DMR generation became a graph, whose nodes are the frames (Action  or Entity frames), while edges are the relations. \ignore{We start describing the DMR generation algorithm by introducing some notations.} Mathematically speaking, we assume that DMR is a graph $\mathcal{G}_d = (\textbf{V},\textbf{E})$, where $\textbf{E}$ is the set of  entity and action frames (\ie the nodes) and $\textbf{E}$ is the set of relations including Case-Role, Domain, and Temporal Relations (\ie the edges).

\begin{algorithm}[h!]
\caption{DMRGen}
\textbf{Input:} $\mathcal{P} = \{p_i, i=1 \to N\}$, Ontology $\mathcal{O}$, Parser CFG Dictionary  $\mathcal{H}$ \\
\textbf{Output:} $\mathcal{G}_d = (\textbf{V},\textbf{E})$ is a graph of Entity and/or Action frames as vertices, where  related frames  are connected by edges\\
\textbf{Initialization:} $\textbf{V}=\emptyset, \textbf{E}=\emptyset$ 
\begin{algorithmic}
\FOR{ all  $p_i$  in $\mathcal{P}$}
\STATE Initialize $\mathcal{E}^i = \emptyset$ and $\mathcal{A}^i  = \emptyset$, where $\mathcal{E}^i$ and  $\mathcal{A}^i$ are the Entity and the Action frames to be detected in $p_i$. 
\STATE Initialize $\mathcal{R}_C^i = \emptyset$,  $\mathcal{R}_D^i  = \emptyset$, and  $\mathcal{R}_T^i = \emptyset$ , where  $\mathcal{R}_C^i$, $\mathcal{R}_D^i$, and $\mathcal{R}_T^i$  are the Case-Role, Domain, and Temporal Relations to be detected in $p_i$. 
\STATE FillFramesAndRelatons($p_i$,  $\mathcal{E}^i$,  $\mathcal{A}^i$, $\mathcal{R}_C^i$, $\mathcal{R}_D^i$, $\mathcal{R}_T^i$, $\mathcal{O}$, $\mathcal{H}$)
\STATE $\textbf{V} = \textbf{V} \cup \mathcal{E}^i \cup \mathcal{A}^i$.
\STATE $\textbf{R} = \textbf{R} \cup \mathcal{R}_C^i \cup \mathcal{R}_D^i \cup \mathcal{R}_T^i$. 
\ENDFOR
\RETURN  $\mathcal{G}_d$
\end{algorithmic}

\begin{algorithmic}
\STATE $\mathbf{function}$  FillFramesAndRelation($p$,  $\mathcal{E}$,  $\mathcal{A}$, $\mathcal{R}_C$, $\mathcal{R}_D$, $\mathcal{R}_T$, $\mathcal{O}$, $\mathcal{H}$)
\STATE fn = GetMatchingRuleFunction($p$, $\mathcal{H}$) 
\IF{$fn$ is not NULL} 
\STATE fn(($p$,  $\mathcal{E}$,  $\mathcal{A}$, $\mathcal{R}_C$, $\mathcal{R}_D$, $\mathcal{R}_T$, Ontology)  
\ELSE 
\FOR{ all  $ch$  in $p.children$}
\STATE FillFramesAndRelations($ch$,  $\mathcal{E}$,  $\mathcal{A}$, $\mathcal{R}_C$, $\mathcal{R}_D$, $\mathcal{R}_T$, Ontology)
\ENDFOR
\ENDIF
\STATE $\mathbf{end}$
\end{algorithmic}
\label{algdmr}
\end{algorithm}
\vspace{-5mm}
The DMR generation procedure is illustrated in algorithm~\ref{algdmr}. The set of frames $\textbf{V}$ and relations $\textbf{E}$ are initialized to empty sets. In order to generate the DMR, each $p_i  \in \mathcal{P}$ is traversed and  the grammatical structure is followed to define subjects, objects, and verbs. Entity frames are created for objects and subjects. While Action Frames are created for verbs. We denote the set of Entity-Frames in the $i^{th}$ sentence by $\mathcal{E}^i = \{ e^i_1,e^i_2 \cdots e^i_{end} \}$, and the set of   Action-Frame in the same sentence as $\mathcal{A}^i = \{ a^i_1,a^i_2 \cdots a^i_{end} \}$. Each frame is augmented with its intended sense from the node information stored in the SEPTs (\ie $n.s$). This step also includes adding attributes of each frame if any. For example, ``small red ball'', a ``ball'' entity frame is generated  with attributes ``small'' and ``red''. The attributes were detected using the parse tree structure. In this case,  ``small'' and ``red'' are adjective of ``ball''. In addition, the type of the attributes is determined by the ontology (\eg red is a color attribute). Simultaneously, the relations (see figure~\ref{fig:MRRelations}) were created following the syntactic structure of the sentences to connect the generated frames. No new frames are created for anaphors, they are linked to the frames they refer exploiting $n.r$.  The newly generated frames for each statement are added to $\textbf{V}$ and they are connected by case-role, domain, and temporal relations following the parse tree structure. These relations are then added to the edges $\textbf{E}$. We name the function that fills all the frames and relation as ``FillFramesAndRelation'', see algorithm~\ref{algdmr}. \ignore{In the $i^{th}$ sentence,  we denote the $j^{th}$ detected case-role, domain relation, temporal relations as ${\textbf{R}_C}_j^i$, ${\textbf{R}_D}_j^i$ and ${\textbf{R}_T}_j^i$. } 

Practically, we implemented the function ``FillFramesAndRelations'' by recursively traversing the given SEPT $p_i$ starting from the root node. Each node is checked  processing each grammar rule in each node in the produced parse trees; This could be possibly any rule that our parser allows. Each rule could be processed differently depending on whether it produces of frames and relations.  For instance, Entity Frames are added to $\mathcal{E}$ while processing $NP$ node (see figure~\ref{fig:MMTPM} Syntactic Analyzer Output).  Another example is that Domain Relations are added to $\mathcal{R}_D$ while processing a rule like $S = DS (``because'',``So that''|``So'') DS$, where there are two statement ($DS$) separated by and preposition that indicate a reason on a result (\eg ``because'',``So that'', and ``So''). Following the same idea, it is not hard to build a function for the rules of interest that to fill the frames and the relations in each statement\footnote{we will put the code of the framework online}. A hash table $\mathcal{H}$ help retrieve each of these function based on the rule as a key.  GetMatchingRuleFunction($p$, $\mathcal{H}$) simply returns the function that processes $p$'s rule  if one exist in $\mathcal{H}$,  and  NULL otherwise. Having traversed all $p_i  \in \mathcal{P}$ by ``FillFramesAndRelations'', a complete graph of all the information in text is created which is the DMR (\ie $\mathcal{G}_d = (\textbf{V},\textbf{E})$). Figure~\ref{fig:MRExample} shows an example meaning representation for the statement "Ali ate the sandwich because he was hungry". \ignore{We assume that the generated graph is connected since it is talking the same topic. If the DMR consists of multiple connected components, the biggest component is only considered. }
\ignore{, is any, using the Ontology by mapping the disambiguated sense of the to its corresponding concept.}

\section{MLMR Generation}
\label{sec:5}
\begin{figure}[h!]
\centering
\includegraphics[width=1.0\textwidth]{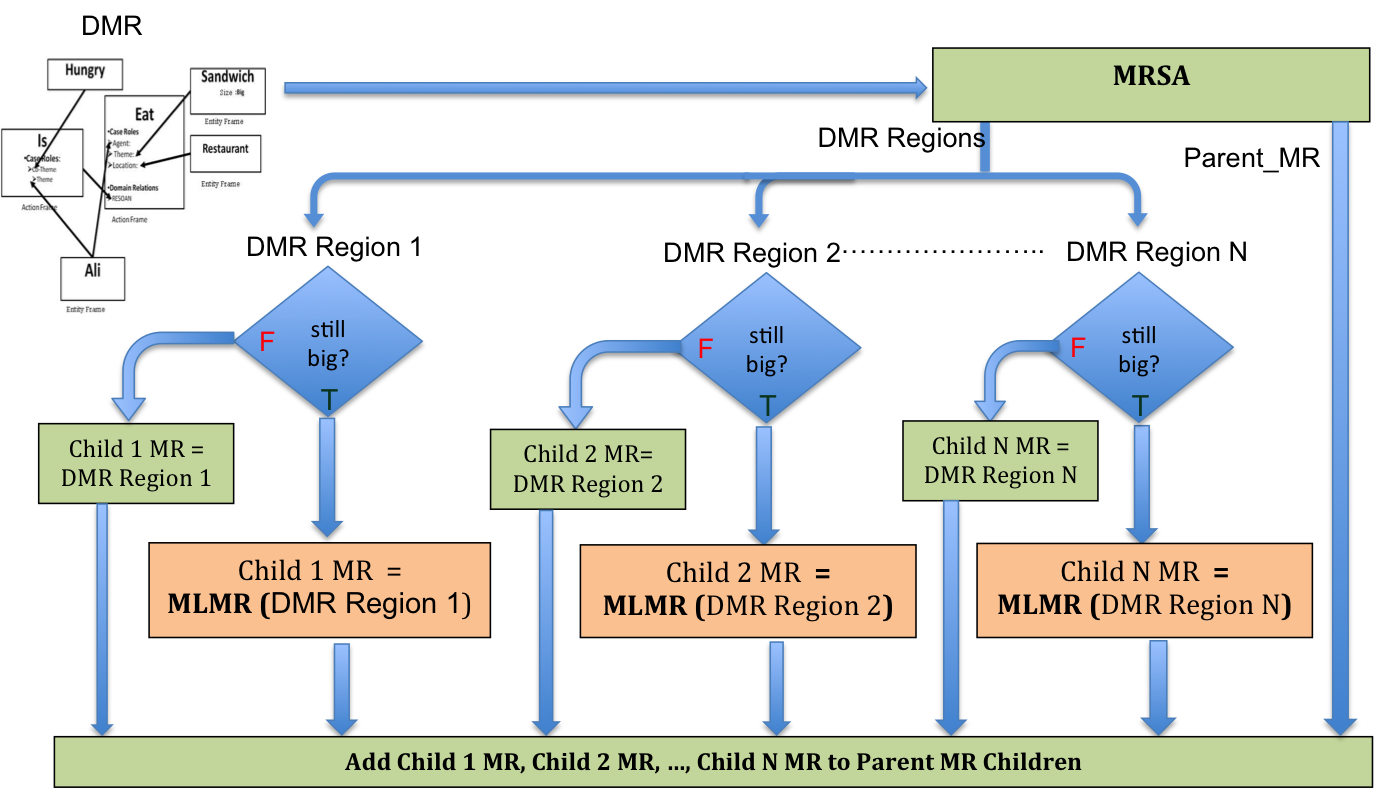}
   \caption{MLMR Recursive Generation (best seen in color). Orange blocks indicates recursion.}
   \label{mlmrpptxfig}
   \vspace{-5mm}
\end{figure}

Figure~\ref{mlmrpptxfig} shows our MLMR recursive generation step. Initially $\mathcal{G}_d$ meaning representation is passed to Meaning Representation Summarization Algorithm (MRSA) algorithm, detailed later in this section\ignore{, to get summarized/abstracted.} Specifically, MRSA generates the abstracted version of   $\mathcal{G}_d$, we name  parent MR $\mathcal{G}_p$,  and it connects parent abstract nodes in the $\mathcal{G}_p$ to regions in the  $\mathcal{G}_d$ that details such abstract nodes in $\mathcal{G}_p$ . We name those regions as DMR Regions. Then each of the DMR Regions are processed recursively by the MLMR process if there are still big to fit in the screen. This termination criterion   could be defined by the number of nodes for instance. In our case, we simply stop after the MR did not change from the previous iteration.  \ignore{We now detail the  MRSA algorithm and and the relevant information in the ontology that we leverage in the algorithm. The main goal of the MRSA  algorithm is to group each set of related actions or entities in the MR into a common concept, so that the information in the text could be represented at an abstract level, we name parent MR. Each frame in the parent MR that semantically group information in the DMR is associated with a sub graph, we name DMR region, which points sub-graph of the DMR that include information that details the frame.}  The remaining part of this section details the MRSA algorithm, the role of the ontology in the algorithm,  and how to make  the MLMR Generation process interactive. 

\subsection{Ontology}

We assume that the ontology has the the property that semantically related concepts have short paths in the hierarchy of the ontology. We also assume that the concepts in the ontology could be mapped to the senses of the frames  stored in the DMR.  We opt to study our framework in one domain to prove that the concept works. Another reason is that natural language research is still progressing towards generalizing  ontologies so that it covers an arbitrary text. The ontology, utilized in this work, consists of 1300 concepts that we created according to the taxonomy of the historical figures topics. The ontology consists of three root nodes of ``Work'', ``Personal Life'', ``Political Life''; the ontology is attached in the supplementary materials\ignore{\footnote{https://sites.google.com/site/mhelhoseiny/projects/natural-language-processing-projects/english2mindmap}}. Concepts presented in the textual information are assumed to be found as sub-nodes under one of these four root nodes. The main reason behind designing our custom ontology is to focus on the evaluation of our main contribution,  Multilevel MindMap visualization, rather than addressing ambiguity problems that exist in the available ontologies (e.g. WordNet \cite{WordNet95}, SUMO \cite{SUMO01}, and CYC \cite{CYC95}), which are out of the scope of this work.

It is important to note that, for each concept in the ontology, we added a field called ``MAP-LEX''. This field basically contains all the senses that matches these concepts. The purpose of this field is to easily map frames in a meaning representation to its corresponding concept. Hence, after the DMR is generated, we augmented each frame in the DMR with the corresponding concept in the ontology by finding the concept, whose ``MAP-LEX'' field contains its stored sense\footnote{All senses are identified by wordNet sense index}. Finally, each concept also contains ``Is-Visual'' flag which indicates whether the concept could be presented as a picture.

\subsection{MRSA}

\begin{figure}[h!]
\vspace{-2mm}
\centering
  \includegraphics[width=0.8\textwidth]{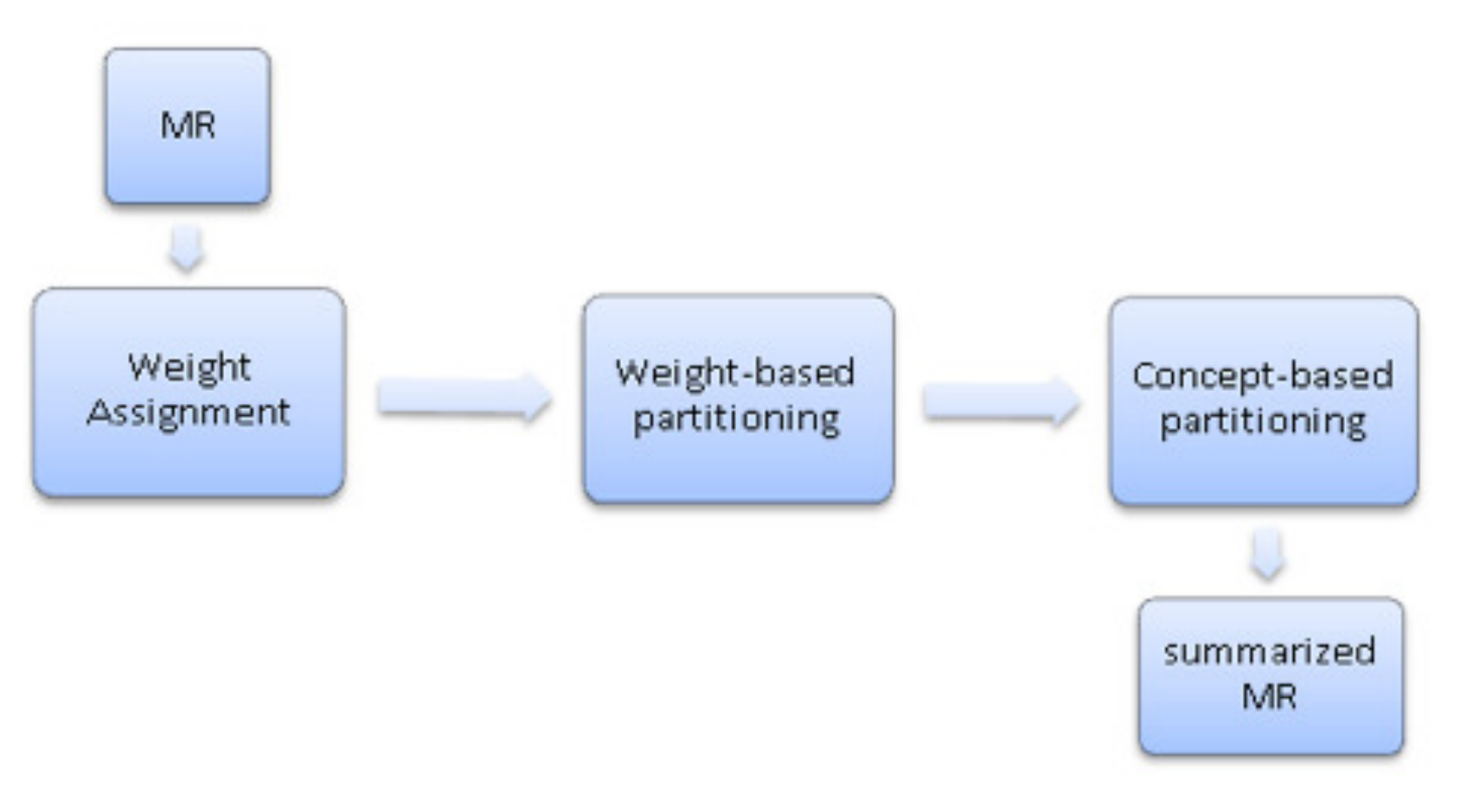}
  \caption{MR Summarization Algorithm (MRSA)}
  \label{fig:MLMR}
\end{figure}

The objective of MRSA is to summarize a meaning representation (MR), based on the ontology and the important frames in the MR. Each frame of the summarized MR is connected to regions (i.e., a set of frames $\textbf{V}_\textbf{c} \subset \textbf{V}$ and relations $\textbf{E}_c \subset  \textbf{E}$) in the input MR that are related to that frame. \ignore{We also denote the summarized MR by parent MR, and each of regions in the input MR that are grouped by MRSA algorithm by child MR.}  Frames in the summarized MR/parent MR that map to more than one frame in the input MR are denoted by \emph{group frames}.  Figure~\ref{fig:MLMR} shows the block diagram of MRSA algorithm. The rationale behind MRSA is to measure the significance of MR frames by a weight assignment phase. It then selects the most important frames (we call them main frames) by clustering, based on the assigned weights. Then, the main frames persist at the top level, with surrounding actions, conceptually grouped according to a conceptual metric (i.e., distance between the concepts inside the ontology hierarchy).  This is achieved by a conceptual based partitioning step. 

\subsubsection{Weight Assignment}
\label{mmwass}
\ignore{The goal of this step is to determine the importance of each frame, in the MR, based on its relations with other frames in the DMR.} In this step, a weight for each frame $f \in V$  of $\mathcal{G}_d$ graph is assigned to indicate its level of significance; the higher the weight, the more significant the frame is.  Each of the case roles,  domain relations, and temporal relations is assigned a constant weight  depending on its type (see figure~\ref{fig:MRRelations}; for example Agent may have a score different from Location or Reason). We observed that important Entity-frames have many important relationships to the other frames. Based on this observation, we model the weight per Entity-Frame as the sum of the weights of its surrounding relations (see Equation \ref{eq:1}).
\begin{equation}\label{eq:1} \small  W_{EF}^k = \sum_{i} N_{CR_i}^kw_{CR_i}+ \sum_{i} N_{DR_i}^kw_{DR_i} \\ +   \sum_{i} N_{TR_i}^kw_{TR_i},  \end{equation} 
where \begin{math}W_{EF}^k\end{math} is the weight of the $k^{th}$ Entity Frame ($\in V$) and  \begin{math}N_{CR_i}^k\end{math} is the number of Case Role relations of type $i$ in of the $k^{th}$ Entity Frame. Similarly, \begin{math} N_{DR_i}^k\end{math} and  \begin{math}N_{TR_i}^k\end{math} are the number of Domain Relations, Temporal Relations of type $i$ in  $f_k$ respectively.      $w_{CR_i}$, $w_{DR_i}$ and $w_{TR_i}$ are the constant weights assigned to CaseRole, Domain and Time Relations of type $i$ (figure~\ref{fig:MRRelations}).  

On the other hand, we assume that an Action-Frame is important if its is connected to important frames. Hence, the weight of each Action-Frame updated by its neighbors (see Equation \ref{eq:2}). 
\begin{equation}
\small
\label{eq:2}
 \begin{split}
       W_{AF}^k =  & \sum_{i} {(R_{CR_i}\sum_{j} w(F_{CR_{ij}}^k)) } + 
         \sum_{i} {(R_{DR_i}  \sum_{j} w(F_{DR_{ij}}^k))} 
        + \sum_{i}{(R_{TR_i} \sum_{j} w(F_{TR_{ij}}^k ))},
     \end{split}
 \end{equation}

\noindent where \begin{math}W_{AF}^k\end{math} is the weight of Action Frame $k$,  $R_{CR_i}$ defines the weight of the case role of type $i$ (figure~\ref{fig:MRRelations}). \begin{math}w(F_{CR_{ij}}^k)\end{math} is  the weight of the $j^{th}$ frame connected to frame $k$ with caserole of type $i$. Similarly, \begin{math} R_{TR_i} \end{math} and \begin{math} R_{DR_i}\end{math}  are the corresponding ratios for temporal and domain relations while \begin{math}w(F_{DR_{ij}}^k) \end{math} and \begin{math} w(F_{TR_{ij}}^k)  \end{math} are the corresponding weights of frames related to frame $k$ with temporal and domain relations respectively. \ignore{If an Action-Frame is surrounded by anther Action-Frame, we initialize the weights of all Action-Frames to zeros. Then, we iteratively update the weights of each Action-Frame by equation~\ref{eq:2} until convergence \footnote{Note that the weights of the  Entity-Frames are computed first and does not change during these iterations}. }

\subsubsection{Weight-based Partitioning}

\ignore{The goal of this step is to identify the main entity frames in the MR.} The entity frame weights, obtained from the weight assignment, are partitioned into clusters using K-Means++ \cite{Arthur:2007} which has an advantage of the initial seeding of the cluster centers. K-Means++ provides a way to avoid producing poor clustering which is more likely to happen by the standard K-Means algorithm. The first cluster center is chosen uniformly at random as one of the data points, then  each subsequent cluster center is chosen from the remaining data points with probability proportional to its squared distance from the point's closest existing cluster center. We used the method in \cite{RayTuri1999} to select the best K with maximum ratio between the intra-cluster distance measure $(\textbf{intra})$ and is the inter-cluster distance measure $(\textbf{inter})$, as illustrated in equation~\ref{eq:kmeansk}.
\begin{equation}
\label{eq:kmeansk}
\small
\begin{split}\textbf{intra}  = \frac 1 N \sum_{i=1}^K \sum_{\textbf{x} \in C_i} {||\textbf{x}-\textbf{z}_i||}^2 , 
\textbf{inter}  = min({||\textbf{z}_i-\textbf{z}_j||^2}),i \neq j 
\end{split} 
\end{equation} 

where $\textbf{z}_i$ and $\textbf{z}_j$ are  centers of clusters $i$ and $j$ respectively, and $C_i$ is the set of points that belong to cluster $i$. Having performed the clustering, we select the cluster of frames of the highest center as the main frames.

\subsubsection{Concept-based Partitioning}

The objective of this step is to group semantically related frames under one common concept according to the  ontology (i.e. abstraction). The associated actions of each main entity frame (extracted from the top cluster of the weight-based partitioning) are passed through the concept-based partitioning and a list of concepts with their corresponding frames is returned. If an action frame of a main entity frame was not grouped with any of the other action frames, the associated entity frames of that action frame are grouped using concept-based partitioning. So, the new summarized meaning representation contains (1) the main entity frames, (2) the grouped concepts associated with them as new action frames, (3) the ungrouped action frames around the main entity frames, and (4) the grouped concepts of entity frames around them as new entity frames. The following steps illustrates how to conceptually partition concepts around each main entity frame.

\begin{enumerate}
      \item Group frames of the exact concept (i.e. ontological distance =0).
      \item Merge two frames whose concepts has the shortest path in the  ontology. 
      \item Repeat step 2, until  the count of the groups $<=$ Gth (We used Gth=6).
\end{enumerate}

\begin{figure}[ht!]
  \vspace{-3mm}
	\center	
	\includegraphics[width=0.77\textwidth] {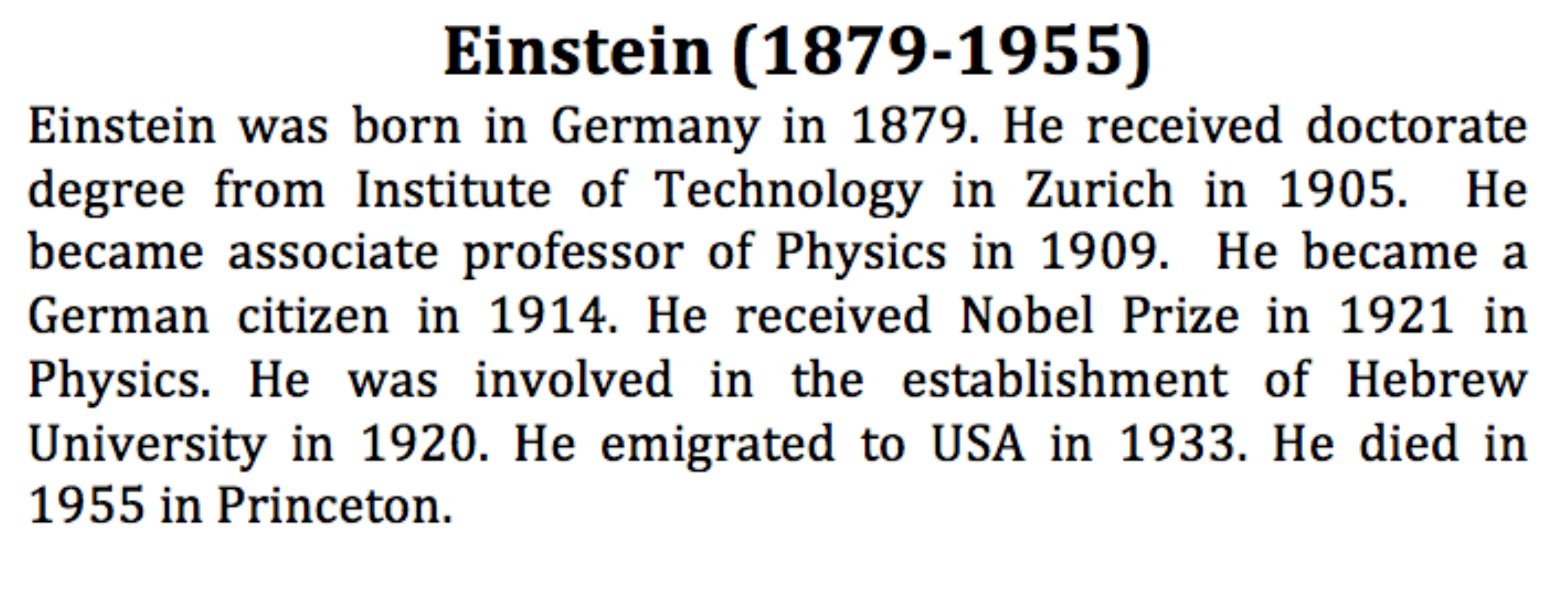}	
 	 \includegraphics[width=0.88\textwidth]{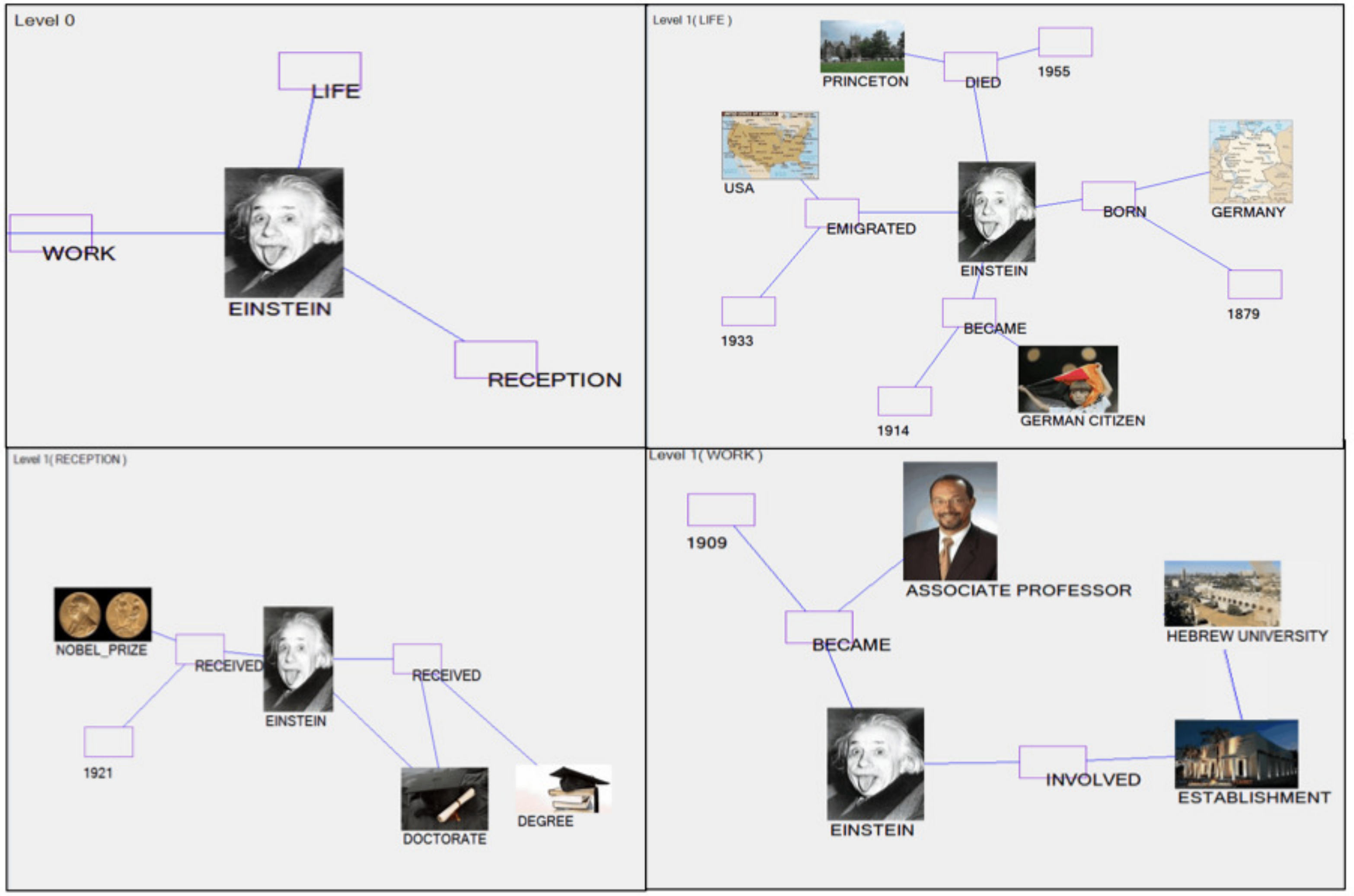}
  \caption{Einstein Multi Level MindMaps}
  \label{fig:EMMG}
  \vspace{-5mm}
\end{figure}

The bottom-up manner of our conceptual grouping algorithm behaves as a variant of Agglomerative hierarchical clustering. The algorithm mainly  analyzes the selected frame(a) and its neighbors and try to do conceptual partitioning if convenient as illustrated in \ignore{figure~\ref{fig:MMG}. for Akhentan example, } figure~\ref{fig:EMMG} for Einstein and figure~\ref{fig:shakessystem} for Shakespeare. The advantage of this model is that each high level frame could be expanded into details (the DMR regions). The highest level of groups serves as the abstracted Meaning representation. the DMR region  for each abstracted frame is generated by including all the frames that are descendants of the given group frame following the Agglomerative hierarchical  clusters. 

In Shakespeare's example, ``Shakespeare'', as a main entity frame, is detected by the weight assignment and partitioning steps. Afterwards, Shakespeare's actions are grouped conceptually using our conceptual-based partitioning through the ontology. For instance, action frames of ``earn'' (living), ``wrote'' (35 good plays), and  ``is'' (great literature) are grouped under ``Work'' concept in the ontology. While,  action frames of ``lived in'' (Strateford), ``born'' (in 1564), and  ``had'' (3 children) are grouped under ``Life/Personal Life'' concept.  From this example, it is intuitive that conceptual partitioning is done in a bottom-up manner, which is adopted in our algorithm. Hence, It takes only 1-2 recursive calls in these examples. In general, the algorithm recursive calls depend on the ontology and the number of concepts in the given instance.

\subsection{Interactive MLMR Generation}
MLMR Generation could be also interactive as follows. Firstly, the DMR is summarized by MRSA.  The summarized MR is then passed to MindMap Generation, detailed in the next section, to pictorially visualize the most abstract level of the MindMap. The user then can interacts with the generated MindMaps by selecting one of the group frames which triggers running MSRA for the subset of DMR referred by the group frame.  The Mindmap generation then visualizes the output MR, which represents the details of the selected group frame. The same process could be repeated interactively until the most detailed level is reached.  Previous steps could be stored to allow going back to parent levels and down to the detailed levels. In this interactive mode, the generation is also effective since only the top level need to be computed, and group frames are detailed based on the interactions by the user. 

\ignore{As introduced in section~\ref{sec:3}, interactive multilevel MindMap visualization is accomplished by repeatedly accessing subsets of the  DMR that are grouped. Firstly, the DMR is summarized by MRSA.  The summarized MR is then passed to MundMap Generation, detailed in the next section, to pictorially visualize the root level of the MindMap. The user then can interacts with the generated MindMaps by selecting one of the group grames which triggers running MSRA for the subset of DMR referred by thr group frame.  The Mindmap generation then visualizes the output MR, which represents the details of the selected group frame. \ignore{ Our solution framework could allow user to even change the grouped information to his preference. Our work is basically an step toward automation the generation of hierarchical MindMap Visualization.}}

\ignore{regionInDMR  is a procedure that retrieves the entities in DMR, which is connected to the selected group frame. This information is stored during the execution of MRSA algorithm. It is easy to see that, MLMR could be also generated without user interaction by iteratively expanding group frames.
}

\section{MindMap Generation}
\label{sec:6}

In the MindMap generation step,  an arbitrary meaning representation is converted into a MindMap that contains images for visual frames and dynamically allocate them on the screen. Visual frames are determined by checking whether the frame's concept is visual in the ontology.  \ignore{After the frame's concept is verified as visual from the ontology, } Then, a  query text is generated for each visual frame to retrieve a relevant image from Google Image Search~\cite{GIS13}. \ignore{ Figure~\ref{fig:shakessystem}  shows an example  single level and multilevel,  automatically generated by our framework}

\vspace{-5mm}
\subsection{Query Generation and Image Web Retrieval}
\label{subsec:61}
\vspace{-3mm}
We propose two approaches to generate the query. In the first approach, the given frame name is associated with its filled attributes. For instance, if there exist an entity frame (Ball: Color= Red, Size=small), the generated query will be ``small red ball''. On the other hand, The second approach involves frame combination. For example, if we have a sentence like ``Shakespeare performed before the queen in December''. If  ``the queen'' is web-searched as suggested by the first approach, an irrelevant image will likely be retrieved. However, if the query combines ``Shakespeare'' and ``queen''. The query is more likely to return a picture of ``Queen Elizabeth'', which is meant by this sentence. That is why we tried to combine frames in the query. We implemented the second approach as follows. Given frame $i$, we add, to the query generated by the first approach, the name of frame(s) within two edges from frame $i$ (i.e., relevant frame). However, we add the names of the frames, only if it is significantly more important. In other words,  relative weight of frame $i$ should be high, with respect to frame $j$. Mathematically speaking,  \begin{math} \frac{WF_i}{WF_j} > th \end{math} where \begin{math} WF_i \end{math} is the weight of frame $i$, \begin{math} WF_j \end{math} is weight of frame $j$ , $th$ is  the relative importance threshold ($th$ =6 in our experiments).

Finally, the images displayed in the generated MindMap are retrieved using Google Image Search (GoIS). We used the GoIS API to retrieve the relevant images using the generated query and the first image in the results is chosen.

\subsection{Automatic Layout allocation}
\ignore{Automatic Layout Allocation is accomplished mainly in response to requirements of data visualization. Therefore, a} A graph drawing algorithm must take into account aesthetics: criteria for making salient characteristics of the graph easily readable. Some aesthetic criteria include minimizing  number of edge crossings, maintaining straight edges as possible,maximize display symmetry, etc. In our work, we achieved an acceptable layout by adopting the spring model by Thomas Kamps et al in \cite{Kamps:1995}.  Spring model meets two aesthetic principles \footnote{Other third party commercial tools/methods could be used also here (\eg Microsoft AGL http://research.microsoft.com/en-us/projects/msagl/}. First, all edges are prefered to have the same length. Second, the layout should be symmetric as possible. The main idea is to model the cost function by attaching a spring between the edges of the desired length as in equation~\ref{eq:Gdlo}.

\begin{equation}
\label{eq:Gdlo}
\begin{split}
E  = &\sum_{i=1}^{n-1} \sum_{j = i+1}^{n} K_{ij}( (x_i-x_j)^2 +(y_i-y_j)^2+l_{ij} ^2-2 l_{ij} \sqrt{(x_i-x_j)^2 +(yi-yj)^2} 
\end{split} 
\end{equation} 

where $(x_i, y_i)$ is the location of the the node $i$.  $l_{ij}$ is the desired length between node $i$  and node $j$ (It is not necessary to have an edge between node $i$ and node $j$). There are four parameters that we designed while adopting the spring model (1) $l$ which is the desired length of directly connected nodes; we set $l$ to  $50$. (2) $D$, which is the diameter of the drawing (i.e. the maximum distance between a pair of nodes in our system); we set $D$ to $600$. (3) $l_{ij} = l$, if there is an edge between node $i$ and node $j$. Otherwise, \ignore{$l_{ij}$ is assigned to the shortest path between node $i$ and node $j$ If node $i$ is not connected to node $j$, then} $l_{ij} = \frac{D \cdot d_{ij}}{min_{l,m} \{ d_{lm} \}}$, where $d_{lm}$ is the length of the shortest path between node $l$ and node $m$. (4) In our system, all-pairs shortest paths are computed , between every pair of nodes, using Floyd Warshall algorithm \cite{Floyd:1962}. $K_{ij}$ is the weight assigned for the cost function between node $i$ and node $j$, which indicate the strength of the spring associate between node $i$ and $j$. $K_{ij} = \frac{1}{{d^*_{ij}}^2}$, where $d^*_{ij} = d_{ij}$ if there is no edge between node $i$ and node $j$, $d^*_{ij} =l$ otherwise. This indicates less restrictive springs, when nodes are not directly connected. \ignore{The parameters of the layout allocation in our system are $l = 50$, $D=600$ pixels}

While nodes are assumed to be points in \cite{Kamps:1995}, each node $i$ in our setting is a rectangular area $R_i$ of width $W_i$ and height $H_i$, centered around the node position $(x_i, y_i)$. For instance, $R_i$ will be  the size of the image to be rendered, if node $i$ is associated with an image. To tackle this problem, we add, to each $l_{ij}$,  $\frac{M_i+M_j}{2}$, where $M_k$ is the diameter of Rectangle $R_k$ of node $k$. The rationale behind adding  $\frac{M_i+M_j}{2}$ is that when the distance between the centers of two rectangles is  $\frac{M_i+M_j}{2}$, the rectangles does not intersect regardless their positions. Finally, the gradient of the cost function is  computed and used to find a local minimum.  We used random re-initialization strategy to reach a good solution minimizing equation~\ref{eq:Gdlo}. We run the spring model 10 times with different initial seeds selected randomly on the screen and the layout with the minimum cost is selected.

\section{Experimental Results}
\label{sec:7}

This section presents the evaluation methodology and experiments of our  MindMap Automation Framework. The evaluation was based on 4900 Mechanical Turk (MTurk)~\cite{mturk11} rating responses, with different system parameters. As indicated in~\cite{mturk12}, MTurk users have less than 2\% of the workers have no high school degree and average education years of $14.9$. In order to improve the quality of the responses,  the workers were exposed to illustrative examples before getting introduced to the instances used for the evaluation. Furthermore, We also required that the MTurk-workers have at least a 95\% approval rate and 1,000 approved assignments \footnote{This is based on their previous history that is maintained in MTurk system} to work on our experiments. The following subsections details the evaluation dataset,  three experiments to evaluate different aspects of our framework.

\subsection{Historical Figures Dataset and Evaluation Metrics}
\ignore{Since this kind of systems have not been comprehensively evaluated, there is no existing datasets to evaluate MindMap Automation. Hence, we created our own dataset to evaluate our  approach.} Since there is no existing datasets to evaluate MindMap Visualization, we created our own dataset to validate our framework. We chose articles of Historical Figures  (e.g., Shakespeare, George Washington, etc) to build our dataset. Thirty-five historical figures were chosen from the BBC historical figures section \cite{BBCHA12}  of size between 150 words to 250 words. We used our framework to generate 455 different MindMaps from these Thirty-five chosen articles. The four hundred and fifty-five MindMaps indicates that thirteen  MindMaps were generated for each article, one for each parameter setting (detailed in this section). The four hundred and fifty-five cases were evaluated based on Human Subject Rankings by 4900 MTurk workers' responses on the following five question survey.

\begin{enumerate}
\item To what extent does the generated output represent the text (Regardless the pictures) ? Grade 1-5.
\item To what extent are the generated pictures  relevant? Grade 1-5. 
\item How many missing actions are there in the shown diagram (if no missing actions, please put 0)?
\item How many missing entities are there in the shown diagram (if no missing  entities , please put 0)?
\item How many repeated entities/actions are there in the shown diagram (if no repeated  entities/actions, please put 0)?
\end{enumerate}

Our evaluation is partitioned into three experiments to evaluate the effect of changing  the system parameters. The framework was evaluated upon three main variations. (a) GoIS Parameters (Size, Image Type)~\cite{GIS13} for single level MindMap in Experiment 1 (b) Concept combination in Experiment 2  (c) Multilevel MindMaps in Experiment 3. The responses of the experiments were evaluated based on three metrics: (1) Mean, (2) Standard Deviation to indicate the stability of the response, and  (3) For grading questions (i.e., Q1,2), Satisfaction Ratio (SR) that is defined as  $ SR = \frac {\# \text{of responses } \ge 4}{\# of responses}$. Appendix A 
shows the generated single-level and multilevel MindMaps for Akhentan, Van Gokh, George Washington, and Newton, a subset of the evaluated outputs ( figures~\ref{fig:MMGAkhentan},~\ref{fig:MMGGokh},~\ref{fig:MMGWash}, and~\ref{fig:MMGNewton} ). More automatically generated MindMaps are attached in the supplementary materials. While, the framework can allow the user to change the retrieved images, we only evaluate results with images automatically retrieved by our method.

\subsection{Experiment 1 (315 cases, 3150 responses )}

In Experiment 1, we  evaluate the MTurk workers' satisfaction with Single Level MindMaps, and  test whether their responses are biased by varying the displayed image parameters.  We tested nine variations: three different image types (All [any image], ClipArt, LineArt) and three sizes (All, Auto, Small). The auto size  is a size mode, we  implemented to determine the image size according to the weight of the frame to be visualized, computed in subsection~\ref{mmwass}. Our implementation uses medium size image ( biggest suitable size for MindMaps) if the number of relations to it is greater than six, otherwise, the system uses small image size. The MindMap generation phase in this experiment did not involve concept combination (which is evaluated in Experiment 2). Overall, we have 315 (9 $\times$ 35) cases with ten MTurk responses each (i.e. 3150 responses).  Table~\ref{tbl:Expr1Q15} (top part) presents the user response statistics grouped by all the nine variations. The results indicate general satisfaction by the MTurk workers.

\begin{table}[ht!]

   \caption{Human Subject Responses for Experiments 1, 2, 3}
   \label{tbl:Expr1Q15}

\centering
  \centering
    \begin{tabular}{|p{1.0cm}|p{0.5cm}|p{0.5cm}|p{0.5cm}|p{0.5cm}|p{0.5cm}|}
   \hline
   \multicolumn{6}{|  c |}{Experiment 1 } \\
    \hline
     &Q1&Q2& Q3 &Q4 & Q5 \\ \hline
    Mean & 4.42&4.34&0.54 & 0.59 & 0.48 \\ \hline
    StdDev & 0.84&0.87&1.58 & 1.74 & 1.68 \\ \hline
    SR & 0.86&0.84& &  &  \\
    \hline
    \hline
   \multicolumn{6}{|c|}{Experiment 2} \\
     \hline
     &Q1    &  Q2   &  Q3 &   Q4    &    Q5    \\ \hline
    Mean &  4.51   &  4.27  &  0.45  &  0.5  &  0.23  \\ \hline
    StdDev &  0.65  &  0.82  &  1.25  &  1.47  &  0.58  \\ \hline
    SR &  0.92  &  0.79  & &  &  \\
    \hline
    \hline
     \multicolumn{6}{|c|}{Experiment 3 } \\
    \hline
        ~        & Q1    & Q2    & Q3    & Q4    & Q5    \\ \hline
        Mean     & 4.48   & 4.33   & 0.59   & 0.61 & 0.25 \\  \hline  
        StdDev & 0.87  & 0.94   & 1.51 & 1.55   & 0.93   \\ \hline
        SR       & 0.89 & 0.84  & ~     &     &  \\  
        \hline
    \end{tabular}
 
\end{table}

\begin{figure}[ht!]
  \centering
  \includegraphics[width=0.8\textwidth,height=2.5in]{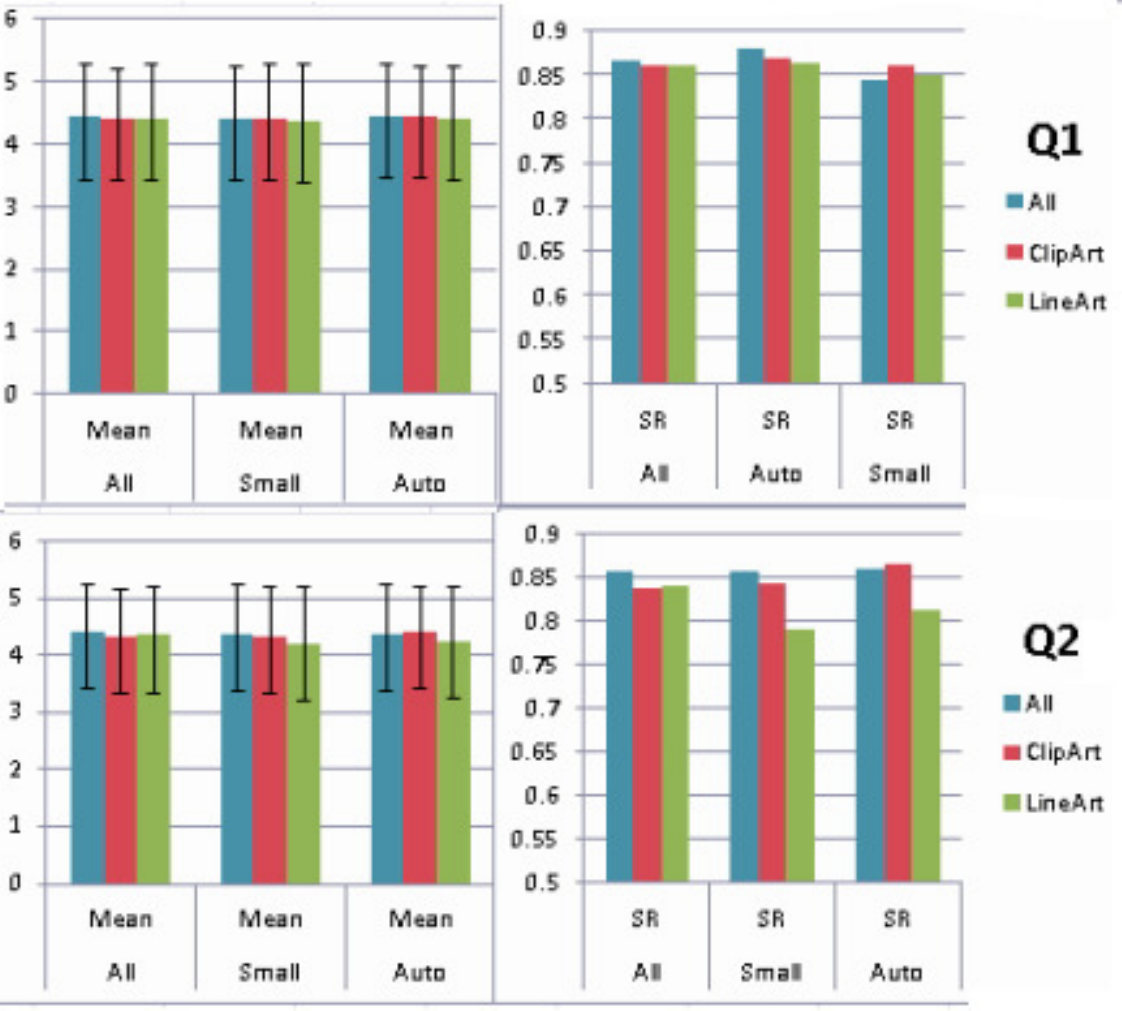}
  \caption{Experiment1 grouped results (Q1,2): Mean and StdDev is on the left while SR is on the right}
  \label{fig:SLResf}
\end{figure}

Q1 and Q2 are the most relevant questions to the appearance of the generated MindMap and the generated images. The results of these questions are grouped for each of the nine variations in this experiment.  Figure~\ref{fig:SLResf}  shows the three evaluation metrics for the nine cases on Q1 and Q2, which indicates that changing the image type or size almost does not affect the responses of the MTurk workers. In other words, these responses were not biased to specific image type or size.

\subsection{Experiment 2 (105 cases,1050 responses )}

Experiment 2 focuses on evaluating the user satisfaction of the MindMaps in case of Concept Combination mode (described in subsection~\ref{subsec:61}). To evaluate whether the MTurk workers were biased to the pictures, we created Concept Combination cases with three variations of image types  resulting in 105 cases (3 $\times$ 35 historical figures). Each case was evaluated by ten different MTurk anonymous workers (1050 responses).

Table~\ref{tbl:Expr1Q15} (middle part) shows summarized results for the experiment. Compared to Experiment 1, there is an improvement of 2\% in the mean and 7\%  in the satisfaction ratio for Q1 with the use of concept combination. However the satisfaction ratio of Q2 (pictures satisfaction) decreased by 10\%. This indicates that the retrieved image using direct query is relatively better than retrieved image using Concept Combination query. The results were also grouped by the three variations of Experiment 2, which indicate a similar behavior to Experiment 1(\ie No bias).

It is important to mention that, the relevance of the MindMap images currently depends only on the quality of the firstly-ranked image retrieved by our generated query from GoIS. After careful look on the MTurk responses and our own evaluation of this experiment, we found that most of the retrieved images are relevant using direct queries. However, in some cases, it does not retrieve a relevant image. For example, "new capital" node in Akhentan example, is assigned to an unsatisfactory image; see figure~\ref{fig:MMGAkhentan} in Appendix A. Our intuition is that context information is not considered in this retrieved images in direct query mode, which we tried to model by  concept combination (CC) queries. Having applied CC mode to Akhentan example,  we managed to retrieve an image of a pharaonic land for this "new capital"  query, which is more relevant. However, the CC mode negatively affected the overall performance of remaining queries since it badly affects most of the remaining frames, as illustrated in the results of this experiment. 

%

\begin{table}[h!]
   \centering
    \begin{tabular}{ | l | l | l | l | l | l | p{1cm} |}
   \hline
   \multicolumn{1}{|c|}{  } & \multicolumn{3}{|c|}{Q1 } &  \multicolumn{3}{|c|}{Q2 } \\ \hline
     &Mean&StdDev& SR &Mean & StdDev&SR \\ \hline
    All &4.56 &0.63 &0.937  &4.31  & 0.825&0.8 \\ \hline
    LineArt &4.53  &0.61 &0.939&4.24  &0.827 &0.76 \\ \hline
    ClipArt &4.46  & 0.71& 0.91&4.26  &0.825 & 0.81\\
    \hline
    \end{tabular}

%
%
%
%
%

     \caption{Experiment 2 (Q1,2) grouped results.}
     \label{tbl:Expr2Q12}
\end{table}

\subsection{ Experiment 3 (35 cases,700 responses)}

The purpose of this experiment is mainly to rate the partitioning of the information in the multilevel MindMaps and not the evaluation of the retrieved images (evaluated in Experiment 1 and 2). Hence, we selected one case for each historical figure (with the best images retrieved in Experiment 1). Q2 was changed to evaluate of conceptual grouping instead of the pictures as they was already evaluated in Experiments 1 and 2. Question 2 became ``To what extent are you satisfied with the structure of the Multi-level MindMap \footnote{Clarification information included in the correct class (e.g Work, Personal Life, and Political Life). As an incorrect instance,  information like the historical figure's  birth or death is classified under Work. This clarification was exposed to the MTurk workers with examples}? Grade 1-5''.   To evaluate it interactively, we have generated 35 interactive flash files  (one for each historical figure). Each flash file was evaluated by twenty MTurk Users (i.e., 700 responses).

Results are shown in Table~\ref{tbl:Expr1Q15} (bottom part). There is 1.4\% and 3\% improvement in the mean and SR respectively compared to the single-level MindMaps.  The results show the satisfaction of the users with the hierarchical representation of the MindMap. While both single-level and multilevel approaches give satisfactory outputs, the multilevel approach is the only way to represent all information in large text, where single level generation is not applicable.

In order to give a careful consideration to the criticism of the generated Multilevel MindMaps, we computed the lowest 15\% for all the  ratings for Q1 and Q2. We also computed the lowest 15\%  percentile for the ratings of each  of the 35 Historical figures and we reported the average overall the historical figures; see table~\ref{tbl:Expr311}.  We further investigated the highest errors reported by the MTurk workers,  in response to Q3, Q4 and Q5, which indicate missing actions, entities, or redundancy of any of them. For each of 35 historical figure, we select the responses that have Non-zero values for Q3, Q4, and Q5. Then, we compute the average over all the historical figures; see table~\ref{tbl:Expr312}.

\begin{table}[h!]
 \caption{Experiment 3 worst 15\% percentile Q1 and Q2: The first row shows the 15\% percentile over the all responses, while, the second row shows the mean of the average of the 15\% for each of historical figure}
   \label{tbl:Expr311}
   \center
\begin{tabular}{|c|c|c|}
\hline 
• & Q1 & Q2 \\ 
\hline 
Mean of 15\% percentile worst responses& 4 & 3 \\ 
\hline 
Mean of the average  15\%  percentile for each 35 historical figures  & 3.78 & 3.57 \\ 
\hline 
\end{tabular} 
\end{table}

\begin{table}[h!]
   \caption{The average of the  non-zero responses' mean of Q3, Q4 and Q5 for each historical figure in  Experiment 3}
   \label{tbl:Expr312}
      \center
\begin{tabular}{|c|c|c|c|}

\hline 
• & Q3 & Q4 & Q5 \\ 
\hline 
Average of Non-zero responses & 3.57 & 2.83 &  2.00\\ 
\hline 
\end{tabular} 
\end{table}

However, this group of responses does not indicate high disagreement with the generated MindMaps as indicated in tables~\ref{tbl:Expr311} and~\ref{tbl:Expr312}, we carefully investigated the provided comments by the human subjects on the three experiments. We noticed that the main criticism of the least satisfied subjects was due to some missing/redundant pieces of information  and  irrelevant images in the MindMap. While, the experiments and our investigation indicate that these cases are few, it is an important feedback to consider for our future work, in which targets generalization on larger scales.

\section{Discussion and Conclusion}
\label{sec:8}
Our framework has multiple limitations. For instance, it depends on the quality of the syntactic analyzer to pass information. If a statement could not be parsed, then it will be ignored, and its information will not be included in the generated MindMap accordingly. Another limitation is that the  MLMR algorithm assumes that there is a central idea of a given document, which is valid for the historical figures dataset, but not valid in general. In the future, we aim at exploring  different methods to generate hierarchy of abstract information in the text and use word vectors  (\eg~\cite{mikolov2013distributed}) to help semantically related information that might not be captured by the ontology. 	We aim to extend the system such that it is reliable in handling very large text\begin{wrapfigure}{r}{0.4\textwidth}   
    \vspace{-7mm}
        \includegraphics[width=0.38\textwidth]{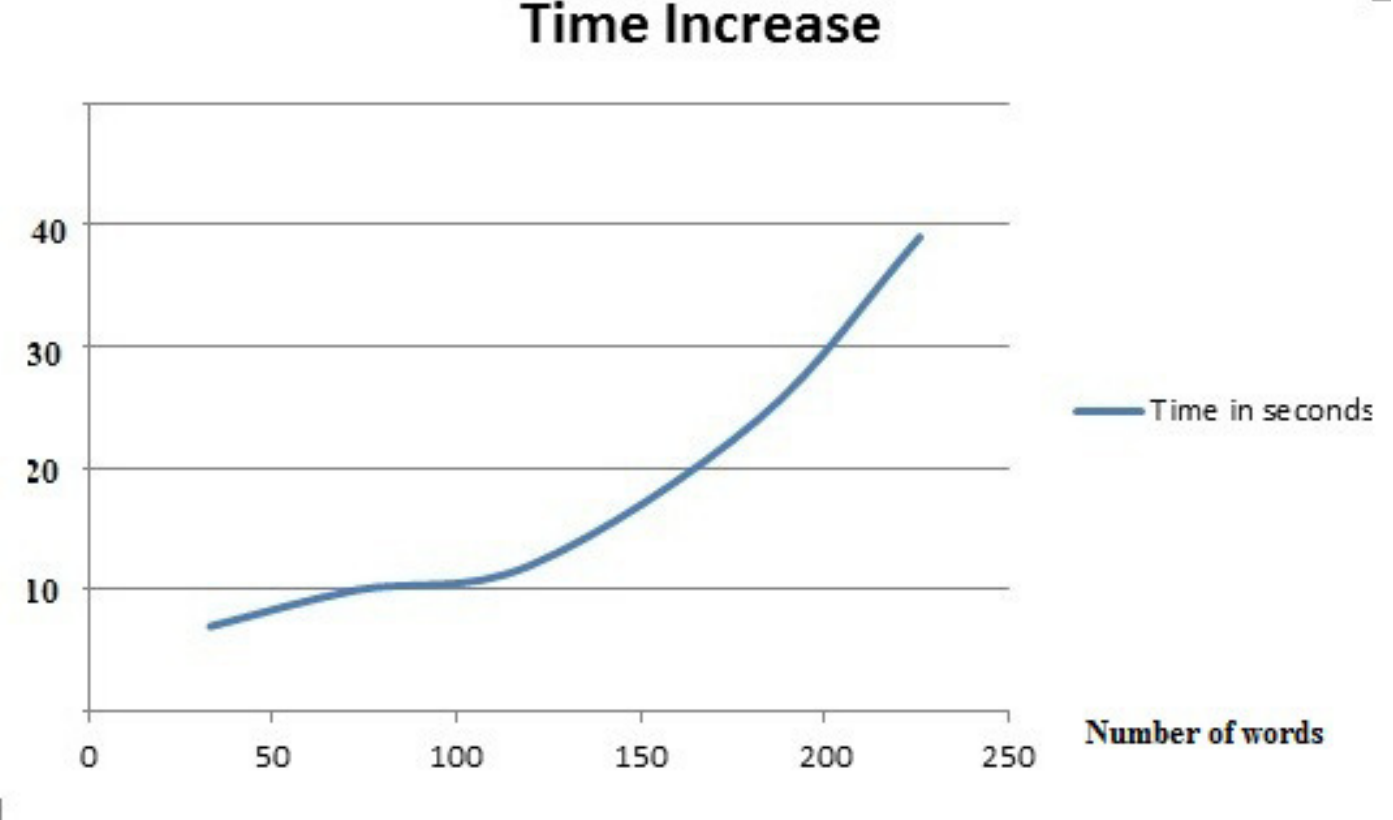}
 \caption{Text-to-Multilevel-MindMap performance on 3.3 GHZ Intel Xeon Processor-6GB Memory}
  \label{fig:SysPf}
    \vspace{-6mm}
\end{wrapfigure} (e.g., a book). Furthermore, we will investigate different approaches to retrieve more relevant images from the textual query. In order to generate MindMaps for a large in a reasonable time,  the framework computational time needs to be significantly reduced. For instance, Multi-level MindMap Generation of a 250 word document take about 40 seconds on a 3.3GHZ-6GB machine; see figure~\ref{fig:SysPf}). The recorded time is the average time to convert raw text to MultiLevel MindMap including automatic layout allocation. Finally, rather than restricting our domain to historical figures in this work, we plan to investigate the  generalization of MindMap visualization for any text and tackle ontology-related  problems.

In conclusion, we have designed and implemented the first automated framework generates a multilevel Mind Map visualization out of English text. The system was comprehensively tested under different parameter settings by MTurk Human Subjects and high satisfaction rates have been recorded. 

\begin{appendices}
\chapter{Appendix A}
\clearpage
\begin{figure}[b]
        \centering           	\includegraphics[width=1.0\textwidth] {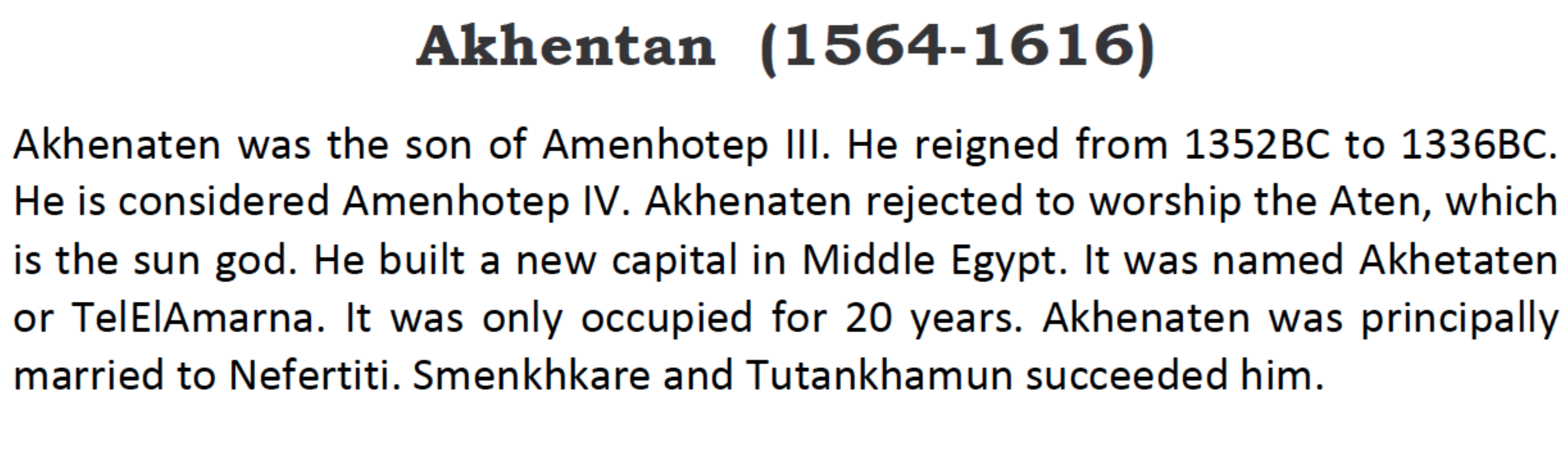}
\includegraphics[width=1.0\textwidth]{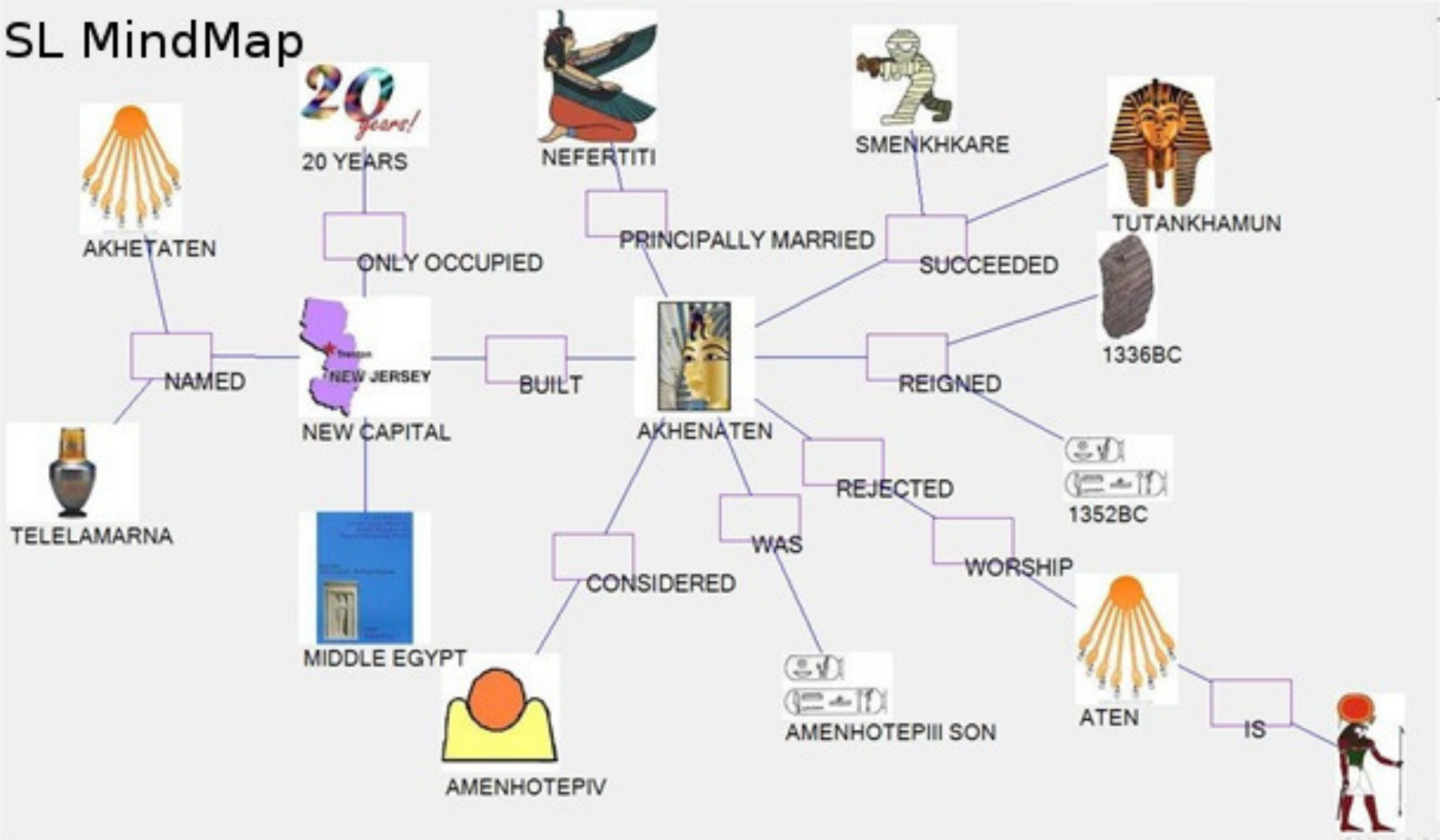}
              	\includegraphics[width=1.0\textwidth]{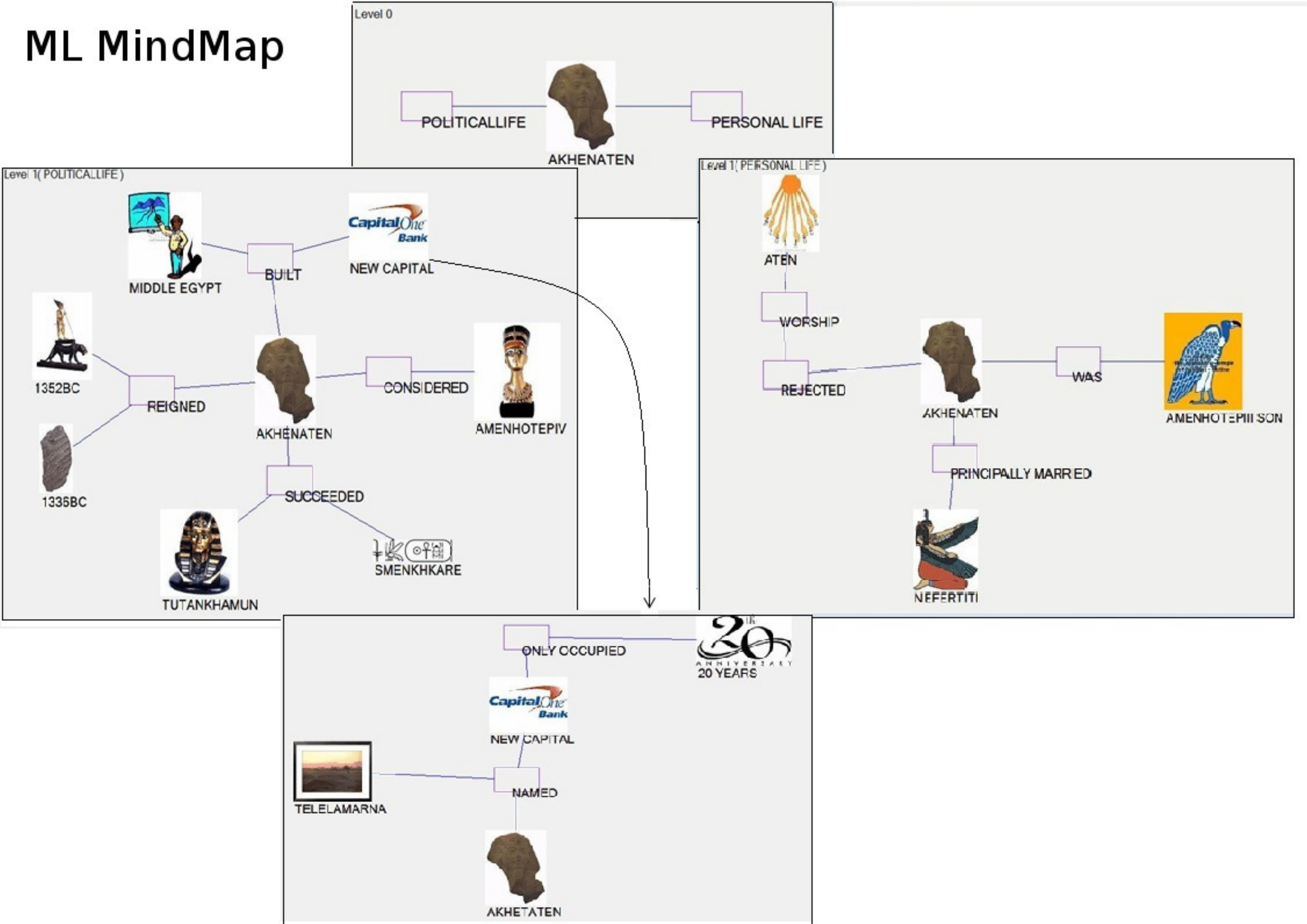}
 \caption{Akhentan Single Level and Multi Level MindMaps}
       \label{fig:MMGAkhentan}
\end{figure}
\begin{figure}
        \centering
         \includegraphics[width=1.0\textwidth] {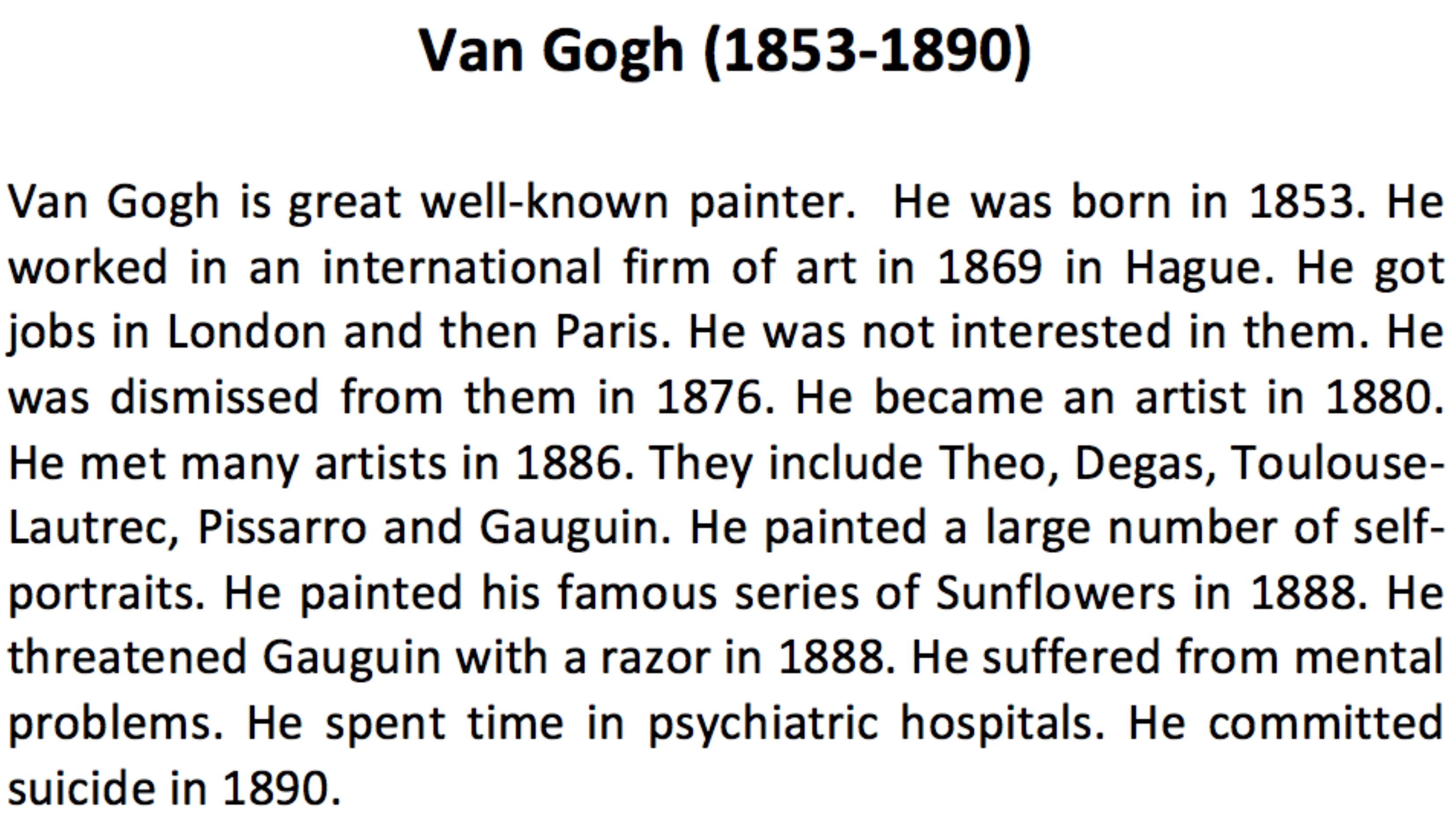}
       	 \includegraphics[width=1.0\textwidth]{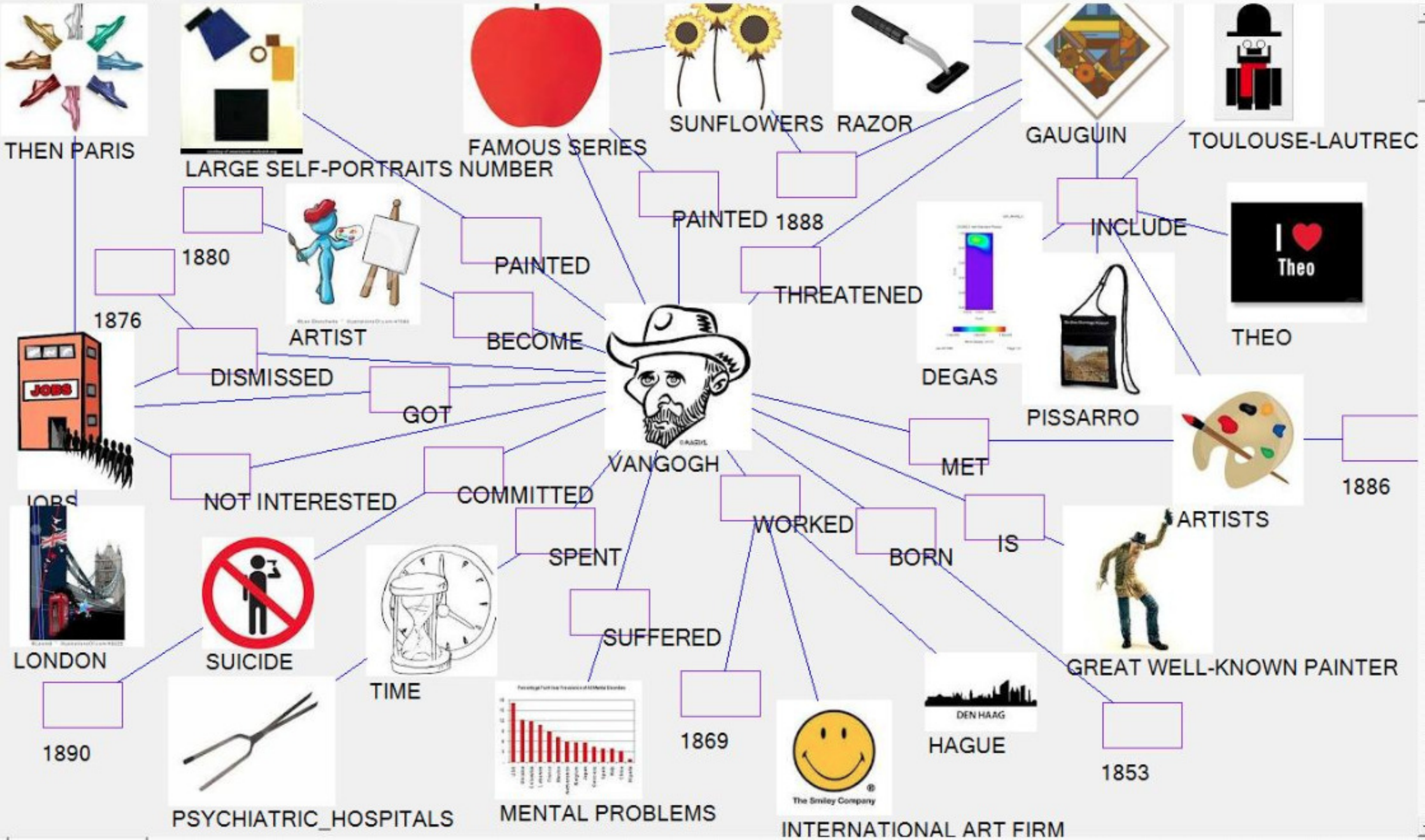}
        	\includegraphics[width=1.0\textwidth]{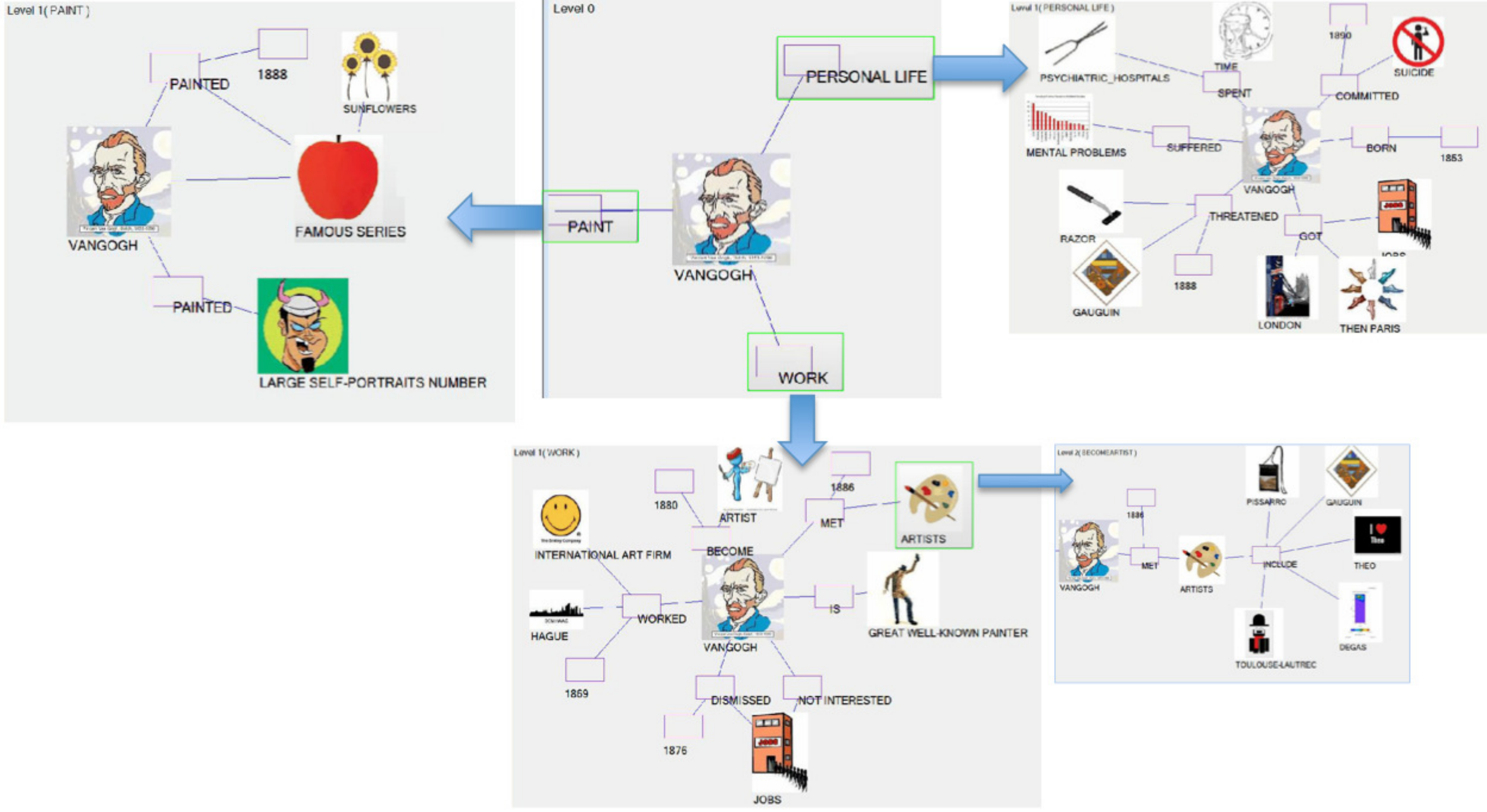}
       \caption{Van Gokh Single Level and Multi Level MindMaps}
  \label{fig:MMGGokh}
\end{figure}

\begin{figure}
        \centering
\includegraphics[width=1.0\textwidth] {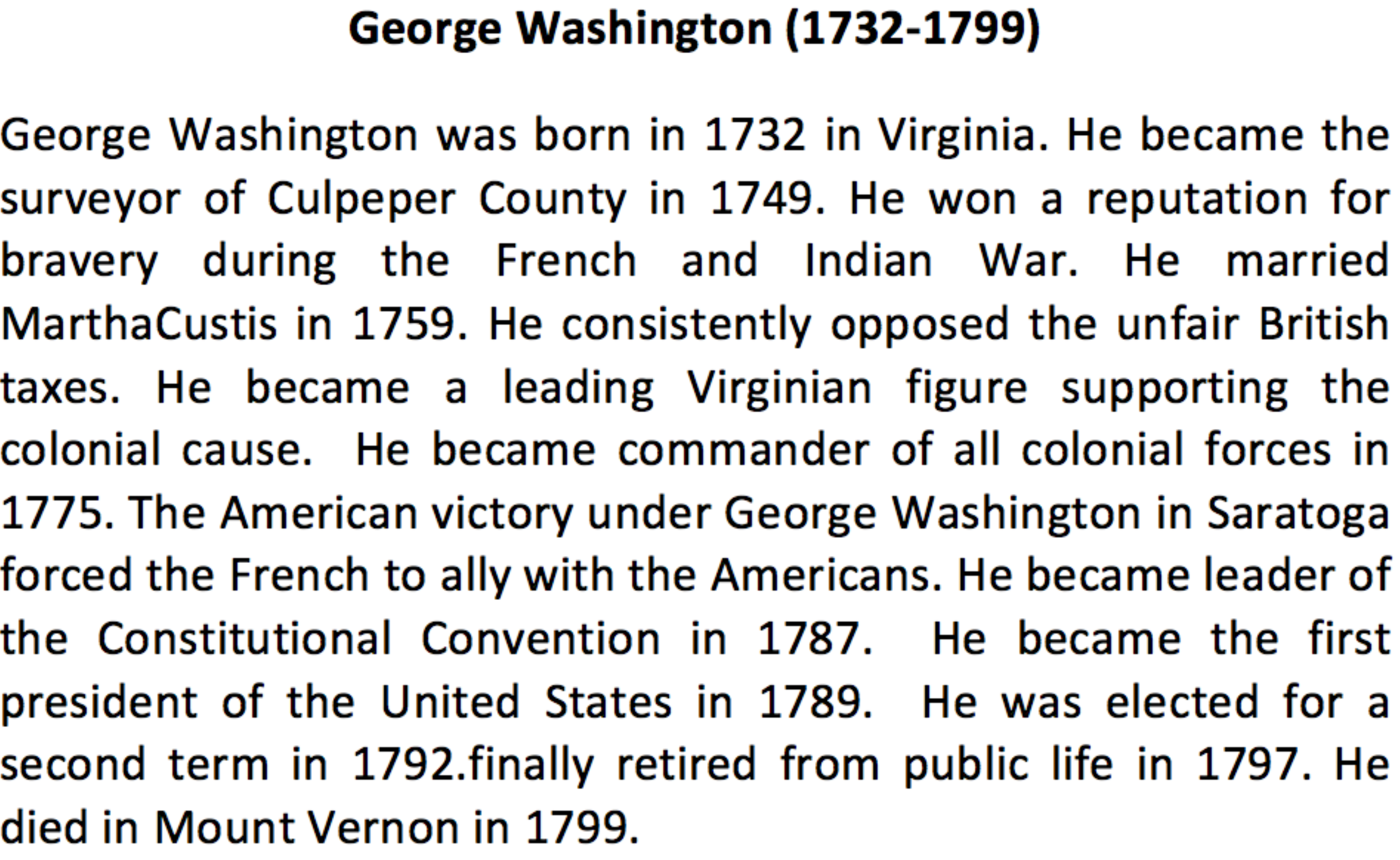}
 	 \includegraphics[width=1.0\textwidth]{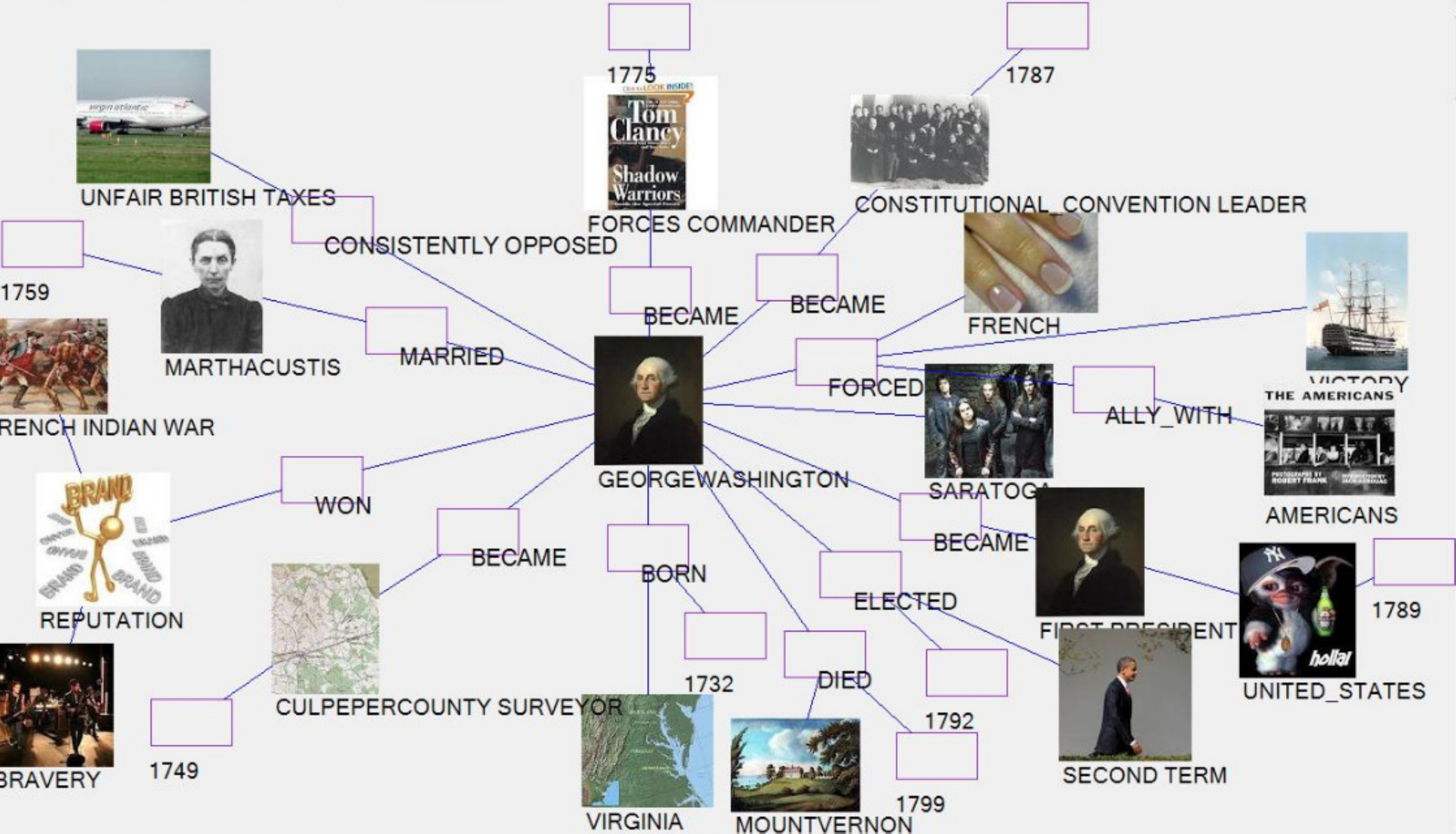}
      	\includegraphics[width=1.0\textwidth]{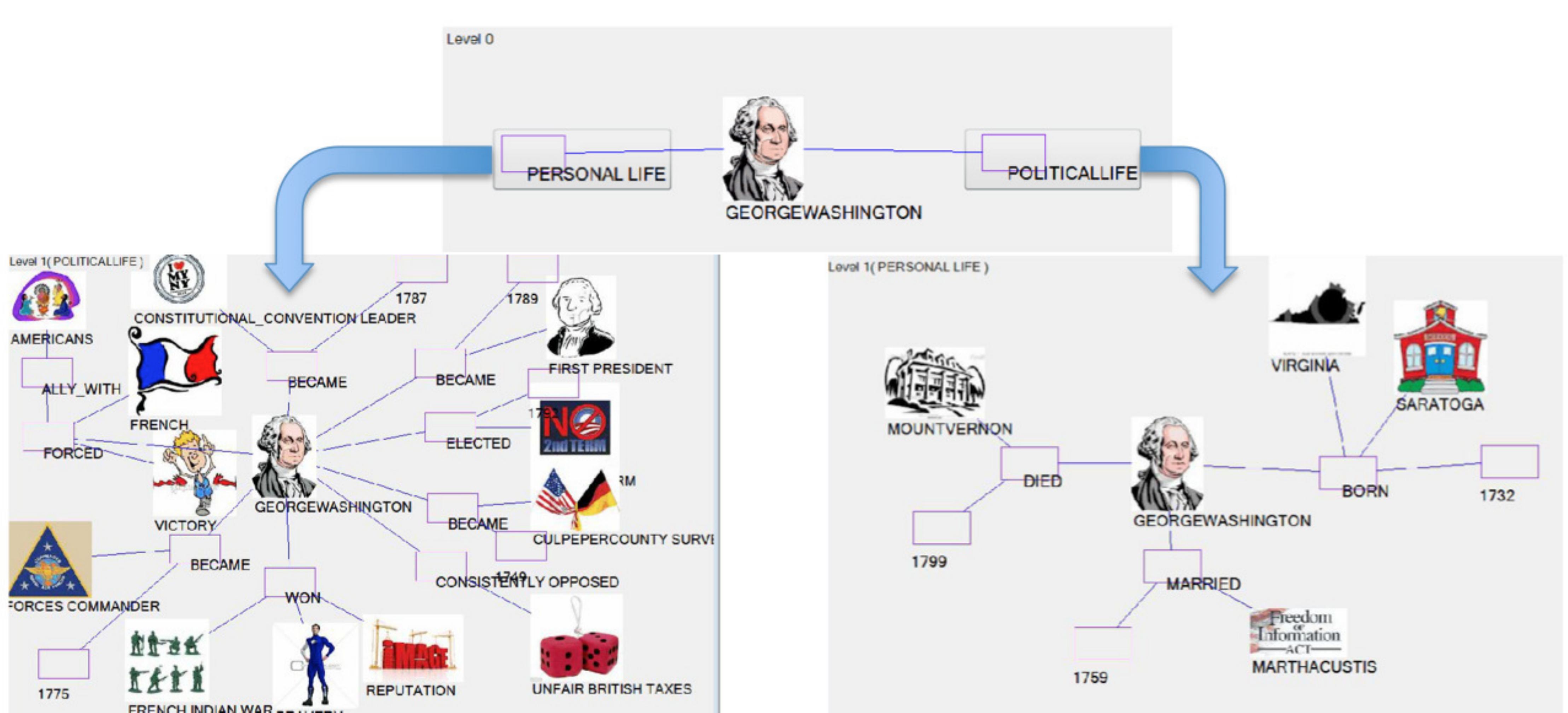}
        \caption{George Washington Single Level and Multi Level MindMaps}
  \label{fig:MMGWash}
\end{figure}

\begin{figure}
        \centering
	 	\includegraphics[width=1.0\textwidth] {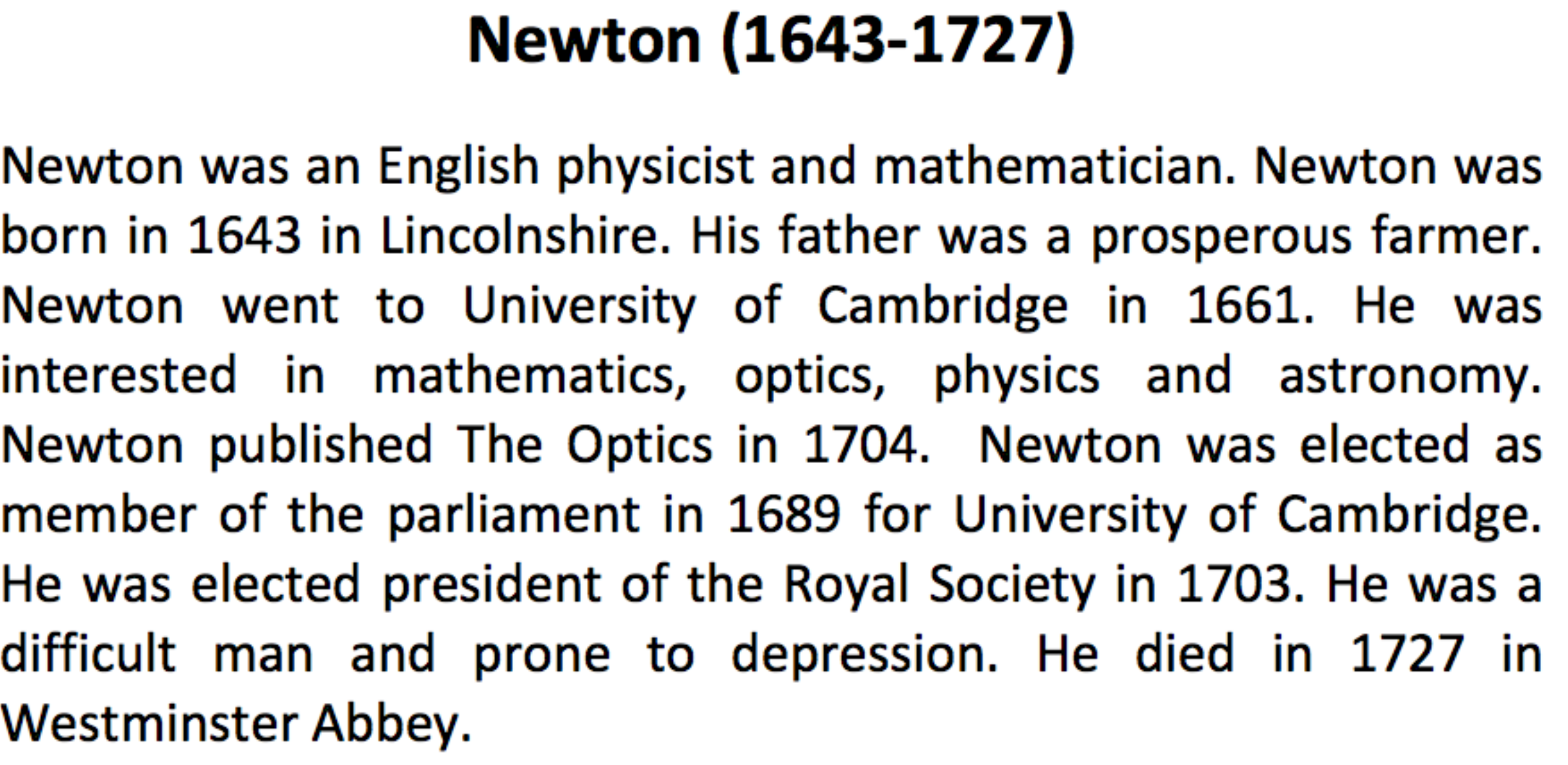}
 \includegraphics[width=1.0\textwidth]{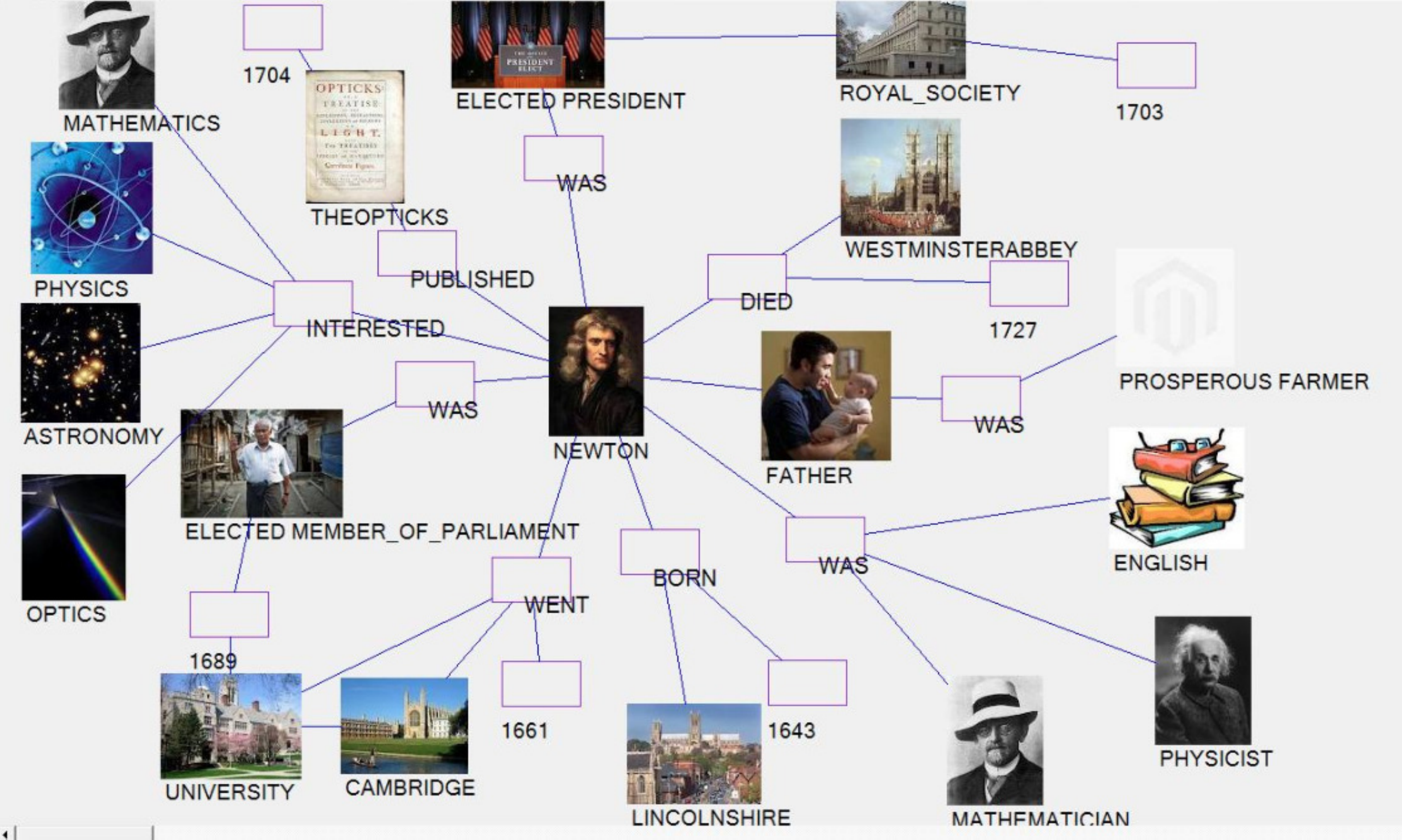}
      \includegraphics[width=1.0\textwidth]{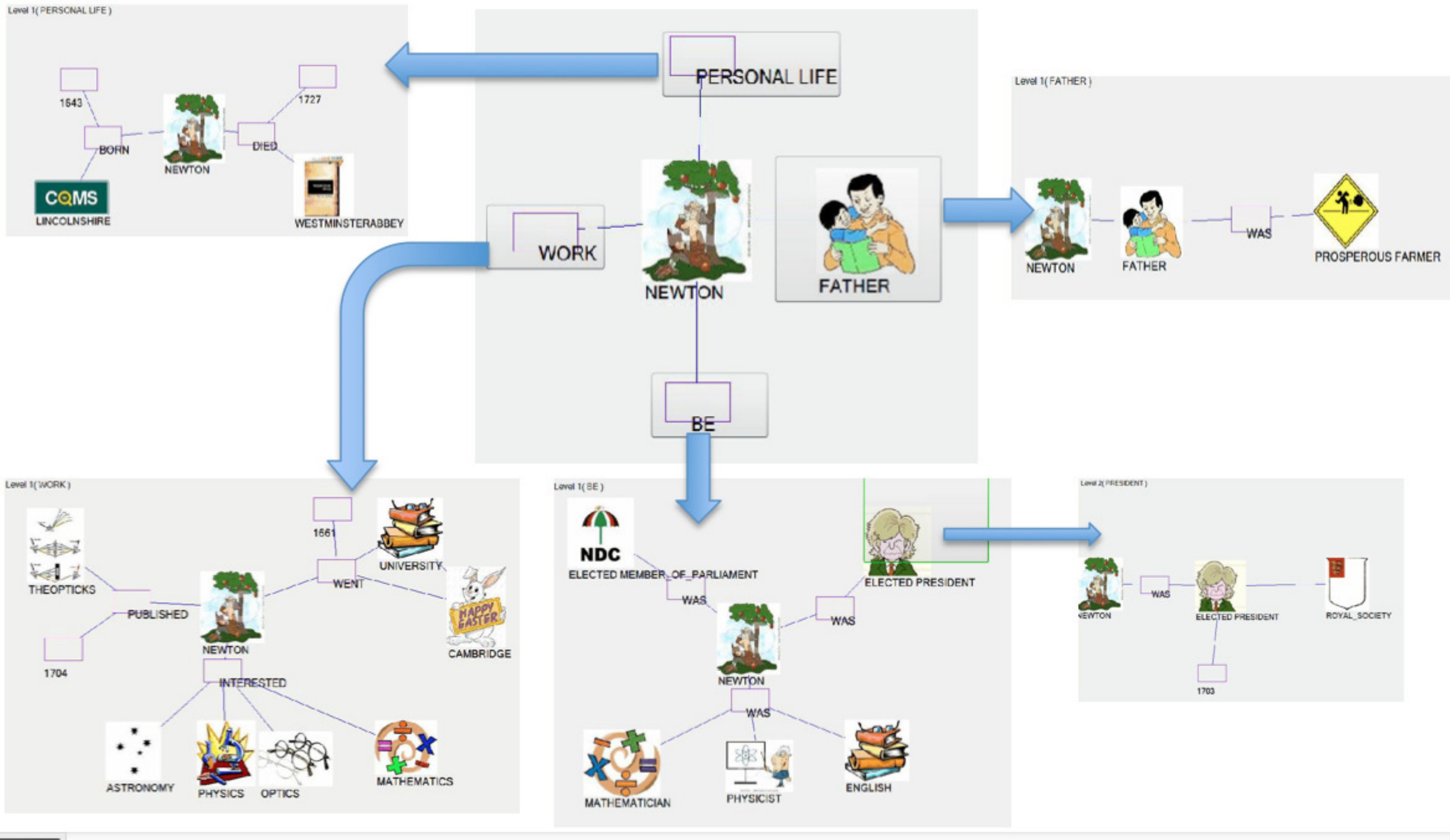}
        \caption{Newton Single Level and Multi Level MindMaps}
  \label{fig:MMGNewton}
\end{figure}
\clearpage
\end{appendices}



\bibliographystyle{spmpsci}      
\bibliography{acmsmall-sample-bibfile}

\begin{thebibliography}{10}
\providecommand{\url}[1]{{#1}}
\providecommand{\urlprefix}{URL }
\expandafter\ifx\csname urlstyle\endcsname\relax
  \providecommand{\doi}[1]{DOI~\discretionary{}{}{}#1}\else
  \providecommand{\doi}{DOI~\discretionary{}{}{}\begingroup
  \urlstyle{rm}\Url}\fi

\bibitem{Afzal_TVCG12}
Afzal, S., Maciejewski, R., Jang, Y., Elmqvist, N., Ebert, D.: Spatial text
  visualization using automatic typographic maps.
\newblock TVCG \textbf{18} (2012)

\bibitem{citeulike:1104130}
Altintas, E., Karsligil, E., Coskun, V.: The Effect of Windowing in Word Sense
  Disambiguation (2005).
\newblock \doi{10.1007/11569596\_65}

\bibitem{Arthur:2007}
Arthur, D., Vassilvitskii, S.: k-means++: the advantages of careful seeding.
\newblock In: Proceedings of the ACM-SIAM symposium on Discrete algorithms.
  Society for Industrial and Applied Mathematics (2007)

\bibitem{Bataineh2009}
B., B., E., B.: An efficient recursive transition network parser for arabic
  language.
\newblock In: World Congress on Engineering (WCE) (2009)

\bibitem{Banerjee02anadapted}
Banerjee, S., Banerjee, S.: An adapted lesk algorithm for word sense
  disambiguation using wordnet.
\newblock In: CICLing (2002)

\bibitem{BBCHA12}
BBC: {BBC Historical Figures}.
\newblock \url{http://www.bbc.co.uk/history/historic_figures/} (2012).
\newblock [Online; accessed January-2012]

\bibitem{mturk12}
Berinsky, A.J., Huber, G.A., Lenz, G.S.: Evaluating online labor markets for
  experimental research: Amazon.com's mechanical turk.
\newblock Political Analysis \textbf{20} (2012)

\bibitem{leftrightEval13}
Berinsky, A.J., Huber, G.A., Lenz, G.S.: An evaluation of the left-brain vs.
  right-brain hypothesis with resting state functional connectivity magnetic
  resonance imaging.
\newblock PLOS ONE \textbf{8} (2013)

\bibitem{mturk11}
Buhrmester, M.D., Kwang, T., Gosling, S.D.: Amazon's mechanical turk: A new
  source of inexpensive, yet high-quality data?
\newblock Perspectives on Psychological Science \textbf{6} (2011)

\bibitem{Collins:2003}
Collins, M.: Head-driven statistical models for natural language parsing.
\newblock Computational Linguistics \textbf{29} (2003)

\bibitem{Conroy:2001}
Conroy, J.M., O'leary, D.P.: Text summarization via hidden markov models.
\newblock In: ACM SIGIR (2001)

\bibitem{Coyne:2010}
Coyne, B., Rambow, O., Hirschberg, J., Sproat, R.: Frame semantics in
  text-to-scene generation.
\newblock In: KES. Springer (2010)

\bibitem{Das_Martins_2007}
Das, D., Martins, A.F.T.: A survey on automatic text summarization.
\newblock Literature Survey for the Language and Statistics II Course at CMU
  (2007)

\bibitem{mmapstudy1}
{Dhindsa}, H.S., {Makarimi-Kasim}, {Roger Anderson}, O.: {Constructivist-Visual
  Mind Map Teaching Approach and the Quality of Students' Cognitive
  Structures}.
\newblock Journal of Science Education and Technology \textbf{20}(2) (2011)

\bibitem{Wenwen_TVCG113}
Dou, W., Yu, L., Wang, X., Ma, Z., Ribarsky, W.: Hierarchicaltopics: Visually
  exploring large text collections using topic hierarchies.
\newblock TVCG \textbf{19} (2013)

\bibitem{mmism12}
Elhoseiny, M., Elgammal, A.: English2mindmap: An automated system for mindmap
  generation from english text.
\newblock In: IEEE International Symposium on Multimedia (ISM) (2012)

\bibitem{mmapstudy2}
Farrand, P., Hussain, F., Hennessy, E.: {The efficacy of the `mind map' study
  technique}.
\newblock Journal of Medical Education \textbf{36} (2002)

\bibitem{Feng_Lapata_2010}
Feng, Y., Lapata, M.: Visual Information in Semantic Representation.
\newblock ACL (2010)

\bibitem{Floyd:1962}
Floyd, R.W.: Algorithm 97: Shortest path.
\newblock Communications of the ACM \textbf{5} (1962)

\bibitem{GIS13}
Google: {Google Image Search}.
\newblock \url{http://www.google.com/advanced_image_search/}.
\newblock [Online; accessed January-2013]

\bibitem{vanHam:2009}
van Ham, F., Wattenberg, M., Viegas, F.B.: Mapping text with phrase nets.
\newblock TVCG \textbf{15} (2009)

\bibitem{MM09}
Hamdy, A., ElHoseiny, M.H., Sahn, R.E., Samier, S., Kamal, E.: Mind maps
  automation (mma) system.
\newblock In: International Conference on Semantic Web and Web Services (SWWS)
  (2009)

\bibitem{ShMMURL:2014}
http://img2.mappio.com/: Shakespear Mindmap (2014).
\newblock
  \urlprefix\url{http://img2.mappio.com/miwisdom/william-shakespeare-short-biography-mind-map-Large.jpg}

\bibitem{txtsummNN04}
Kaikhah, K.: Automatic text summarization with neural networks.
\newblock In: International IEEE Conference on Intelligent Systems (2004)

\bibitem{Kamps:1995}
Kamps, T., Kleinz, J., Read, J.: Constraint-based spring-model algorithm for
  graph layout.
\newblock In: Proceedings of the Symposium on Graph Drawing (1996)

\bibitem{MMMobile11}
Kudelic, R., Konecki, M., Malekovic, M.: Mind map generator software model with
  text mining algorithm.
\newblock In: International Conference on Information Technology Interfaces
  (ICITI) (2011)

\bibitem{Lappin:1990}
Lappin, S., McCord, M.: A syntactic filter on pronominal anaphora for slot
  grammar.
\newblock In: ACL (1990)

\bibitem{Leass94analgorithm}
Leass, H.J.: An algorithm for pronominal anaphora resolution.
\newblock Computational Linguistics \textbf{20} (1994)

\bibitem{CYC95}
Lenat, D.: Cyc: A large-scale investment in knowledge infrastructure.
\newblock Communications of the ACM \textbf{38} (1995)

\bibitem{Litkowski_2001}
Litkowski, K.C.: Syntactic Clues and Lexical Resources in Question-Answering
  (2001)

\bibitem{Ma06}
Ma, M.: Automatic conversion of natural language to 3d animation.
\newblock Ph.D. thesis, University of Ulster (2006)

\bibitem{TMap2006}
Maicher, L., Park, J. (eds.): Charting the Topic Maps Research and Applications
  Landscape.
\newblock Springer (2006)

\bibitem{Marcu:1998}
Marcu, D.C.: The rhetorical parsing, summarization, and generation of natural
  language texts.
\newblock {PhD} dissertation (1998)

\bibitem{stanford_dependencies}
de~Marneffe, M.C., MacCartney, B., Manning, C.D.: Generating typed dependency
  parses from phrase structure trees.
\newblock In: LREC (2006)

\bibitem{mikolov2013distributed}
Mikolov, T., Sutskever, I., Chen, K., Corrado, G.S., Dean, J.: Distributed
  representations of words and phrases and their compositionality.
\newblock In: Advances in Neural Information Processing Systems, pp. 3111--3119
  (2013)

\bibitem{WordNet95}
Miller, G.A.: Wordnet: A lexical database for english.
\newblock Communications of the ACM \textbf{38} (1995)

\bibitem{WSDSurvey09}
Navigli, R.: Word sense disambiguation: A survey.
\newblock ACM Computing Surveys \textbf{41} (2009)

\bibitem{SUMO01}
Niles, I., Pease, A.: Towards a standard upper ontology.
\newblock In: FOIS (2001)

\bibitem{Novak06thetheory}
Novak, J.D.: The theory underlying concept maps and how to construct them.
\newblock Tech. rep. (2006)

\bibitem{IMindTool:2012}
novamind.com: IMind tool (2012 (accessed February, 2012)).
\newblock \urlprefix\url{http://www.novamind.com/}

\bibitem{Osborne:2002}
Osborne, M.: Using maximum entropy for sentence extraction.
\newblock In: ACL Workshop on Automatic Summarization (2002)

\bibitem{TMap2003}
Park, J., Hunting, S. (eds.): XML Topic Maps: Creating and Using Topic Maps for
  the Web.
\newblock Addison-Wesley Longman Publishing Co., Inc., Boston, MA, USA (2003)

\bibitem{Qiu04apublic}
Qiu, L., yen Kan, M., seng Chua, T.: A public reference implementation of the
  rap anaphora resolution algorithm.
\newblock In: LREC (2004)

\bibitem{RayTuri1999}
Ray, S., Turi, R.H.: Determination of number of clusters in k-means clustering
  and application in colour image segmentation.
\newblock ICAPRDT  (1999)

\bibitem{NovaMindTool:2012}
thinkbuzan.com: IMind tool (2012 (accessed February, 2012)).
\newblock
  \urlprefix\url{http://www.thinkbuzan.com/us/?utm_nooverride=1&gclid=CPaV7saVkK4CFUff4AodtTRidg}

\bibitem{mmapbook:buzan}
Tony~Buzan, B.B., Harrison, J.: The Mind Map Book: Unlock Your Creativity,
  Boost Your Memory, Change Your Life.
\newblock Pearson Education Ltd (2010)

\bibitem{Viegas09}
Viegas, F., Wattenberg, M., Feinberg, J.: Participatory visualization with
  wordle.
\newblock TVCG \textbf{15} (2009)

\bibitem{Zhang2009}
Youzhi, Z.: Research and implementation of part-of-speech tagging based on
  hidden markov model.
\newblock In: Asia-Pacific Conference on Computational Intelligence and
  Industrial Applications (2009)

\end{thebibliography}

\begin{wrapfigure}{l}{0.2\textwidth}
\vspace{-8mm}   
     \includegraphics[width=0.2\textwidth]{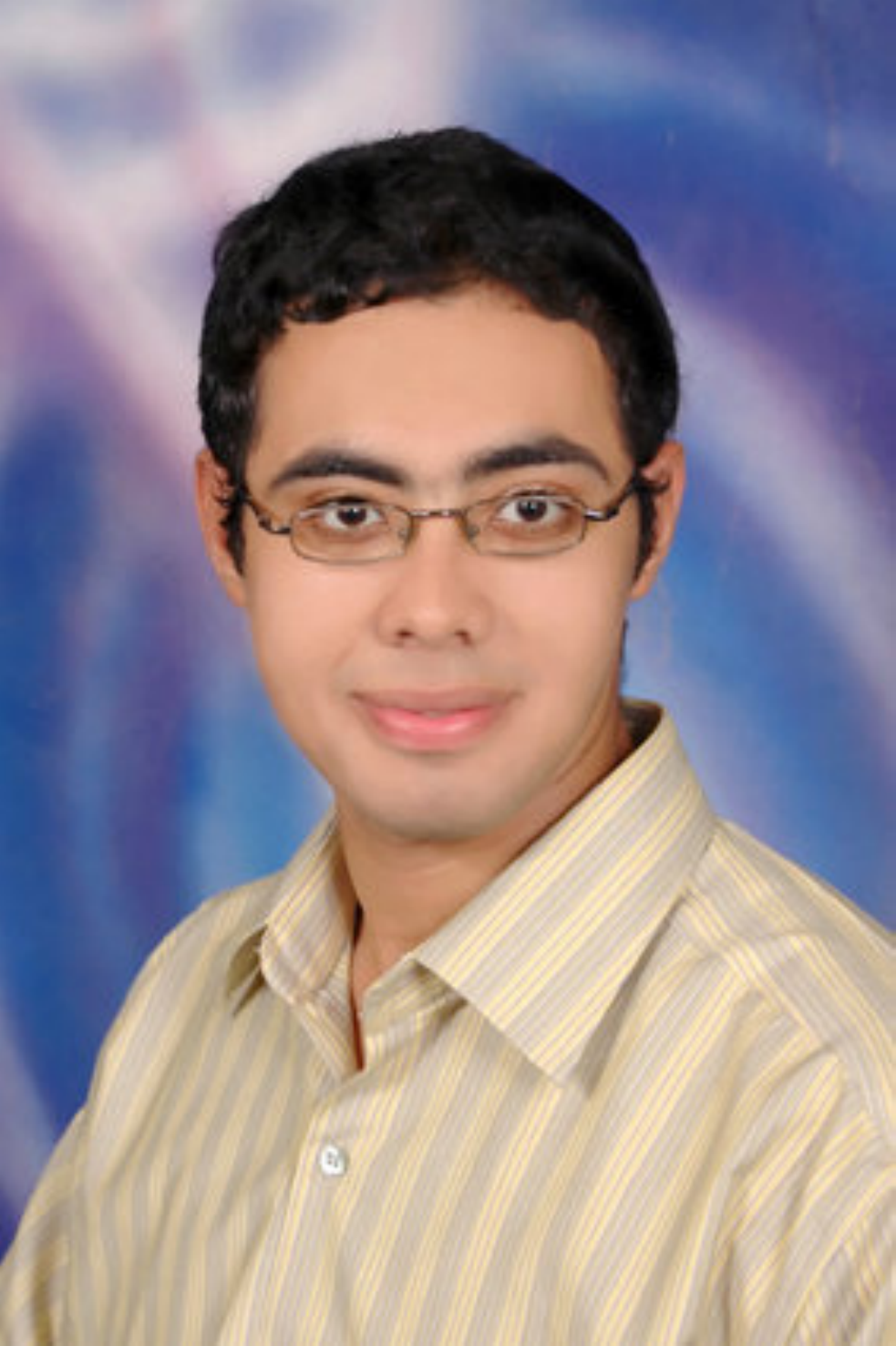}
     \vspace{-8mm} 
\end{wrapfigure}
\textbf{Mohamed Elhoseiny }is a PhD student at the Department of Computer Science, Rutgers, the State University of New Jersey. He is also a member of the Center for Computational Biomedicine Imaging and Modeling (CBIM).  He received his B.Sc. and M.Sc. degrees in computer science from University of AinShams, Egypt in 2006 and 2010, respectively. He is currently pursuing his studies in computer vision and multimedia as a PhD student in Computer Science department at Rutgers University.
\vspace{2mm}
 \begin{wrapfigure}{l}{0.2\textwidth}  
\vspace{-5mm} 
         \includegraphics[width=0.2\textwidth]{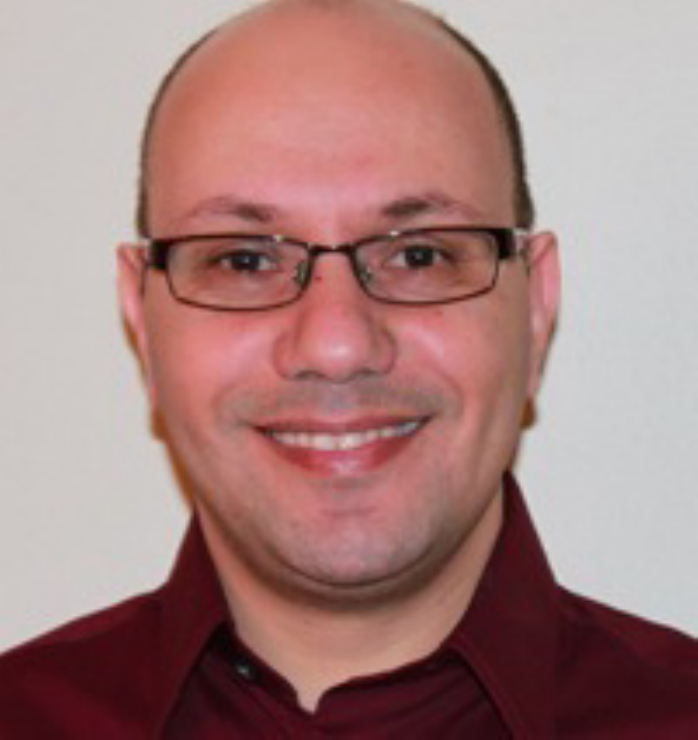}
         \vspace{-10mm} 
\end{wrapfigure}\,\,\;

 \textbf{Ahmed Elgammal} is an associate professor at the Department of Computer Science, Rutgers, the State University of New Jersey Since Fall 2002. Dr. Elgammal is also a member of the Center for Computational Biomedicine Imaging and Modeling (CBIM). His primary research interest is computer vision and machine learning. His research focus includes human activity recognition, human motion analysis, tracking, human identification, and statistical methods for computer vision. Dr. Elgammal received the National Science Foundation CAREER Award in 2006. Dr. Elgammal has been the Principal Investigator and Co-Principal Investigator of several research projects in the areas of Human Motion Analysis, Gait Analysis, Tracking, Facial Expression Analysis and Scene Modeling; funded by NSF and ONR. Dr. Elgammal is Member of the review committee/board in several of the top conferences and journals in the computer vision field. Dr. Elgammal received his Ph.D. in 2002 from the University of Maryland, College Park. He is a senior IEEE member.

\end{document}